\documentclass[11pt]{article}
\usepackage[margin=1.2in]{geometry}
\usepackage{fontspec}
\usepackage{natbib}
\usepackage{microtype}
\usepackage{titlesec}
\usepackage{hyperref}
\usepackage{amsmath,amssymb}

\titleformat*{\section}{\normalfont\bfseries\large}
\titleformat*{\subsection}{\normalfont\bfseries\normalsize}

\titlespacing*{\section}{0pt}{1.7em}{0.8em}
\usepackage{xcolor}
\definecolor{linkblue}{RGB}{83, 22, 204}

\hypersetup{
  colorlinks=true,
  linkcolor=linkblue,
  citecolor=linkblue,
  urlcolor=linkblue,
  pdfborder={0 0 0}
}

\usepackage{fontspec}
\usepackage{titlesec}
\titleformat{\paragraph}[runin]
  {\normalfont\bfseries}
  {}{0pt}{}

\titlespacing*{\paragraph}
  {0pt}{1em}{1em}

\usepackage{hyperref}
\usepackage{url}

\usepackage{graphicx}
\usepackage{hyperref}
\usepackage{textcomp}
\usepackage{url}
\usepackage{amsthm}
\usepackage{amssymb}
\usepackage{booktabs}
\usepackage{listings}
\usepackage{float}
\usepackage{wrapfig}
\usepackage{lipsum}
\usepackage{subcaption}
\newtheorem{theorem}{Theorem}

\newtheorem{proposition}[theorem]{Proposition}

\newtheorem{definition}[theorem]{Definition}

\usepackage{graphicx}
\usepackage{hyperref}
\usepackage{url}
\usepackage{amsthm}
\usepackage{amssymb}
\usepackage{fvextra}
\usepackage{algorithm}
\usepackage{algpseudocode}
\usepackage{changepage}
\usepackage{listings}
\usepackage[most]{tcolorbox}
\tcbuselibrary{listings,breakable,skins}

\DefineVerbatimEnvironment{ExampleBlock}{Verbatim}{
  fontsize=\scriptsize,
  breaklines=true,
  breakanywhere=true,
  frame=single,
  framesep=2mm
}

\usepackage{listings}
\usepackage[most]{tcolorbox}
\tcbuselibrary{listings,breakable}

\usepackage{listings}
\usepackage[most]{tcolorbox}

\tcbuselibrary{listings,breakable,skins}

\newcommand{\exampleheading}[1]{%
  \par\medskip
  \noindent\hspace*{0em}\textbf{#1}%
  \par\nobreak\vspace{8pt}%
}

\input{theorem_boxing_style.tex}

\usepackage{amsmath,amsfonts,bm}

\def\eqref#1{equation~\ref{#1}}

\def\1{\bm{1}}

\def\rvs{{\mathbf{s}}}

\def\vs{{\bm{s}}}

\DeclareMathAlphabet{\mathsfit}{\encodingdefault}{\sfdefault}{m}{sl}
\SetMathAlphabet{\mathsfit}{bold}{\encodingdefault}{\sfdefault}{bx}{n}

\usepackage[most]{tcolorbox}
\tcbuselibrary{listings,breakable,skins}
\tcbset{
  modeltranscript/.style={
    enhanced,
    listing only,
    breakable,
    colback=white,
    colframe=black!25,
    boxrule=0.4pt,
    sharp corners,
    boxsep=0pt,
    left=2.5mm,
    right=2.5mm,
    top=2mm,
    bottom=2mm,
    before skip=10pt,
    after skip=12pt,
    colbacktitle=white,
    coltitle=black,
    fonttitle=\normalfont\footnotesize,
    toptitle=2.5mm,
    bottomtitle=2.5mm,
    lefttitle=2.5mm,
    righttitle=2.5mm,
    titlerule=0.4pt,
    listing options={
      basicstyle=\ttfamily\scriptsize,
      breaklines=true,
      breakatwhitespace=false,
      breakautoindent=false,
      breakindent=0pt,
      columns=fullflexible,
      keepspaces=true,
      showstringspaces=false,
      tabsize=2,
      literate=*%
        {A}{{A\allowbreak}}1%
        {C}{{C\allowbreak}}1%
        {G}{{G\allowbreak}}1%
        {T}{{T\allowbreak}}1%
        {N}{{N\allowbreak}}1%
        {Å}{{\AA}}1%
        {°}{{\ensuremath{^\circ}}}1%
        {±}{{\ensuremath{\pm}}}1%
        {²}{{\ensuremath{^2}}}1%
        {³}{{\ensuremath{^3}}}1%
        {·}{{\ensuremath{\cdot}}}1%
        {×}{{\ensuremath{\times}}}1%
        {α}{{\ensuremath{\alpha}}}1%
        {β}{{\ensuremath{\beta}}}1%
        {γ}{{\ensuremath{\gamma}}}1%
        {–}{{--}}1%
        {—}{{---}}1%
        {…}{{\ldots}}1%
        {⁵}{{\ensuremath{^5}}}1%
        {⁶}{{\ensuremath{^6}}}1%
        {⁺}{{\ensuremath{^+}}}1%
        {⁻}{{\ensuremath{^-}}}1%
        {₂}{{\ensuremath{_2}}}1%
        {₃}{{\ensuremath{_3}}}1%
        {₄}{{\ensuremath{_4}}}1%
        {₆}{{\ensuremath{_6}}}1%
        {₉}{{\ensuremath{_9}}}1%
        {→}{{\ensuremath{\rightarrow}}}1%
        {↔}{{\ensuremath{\leftrightarrow}}}1%
        {−}{{\ensuremath{-}}}1%
        {√}{{\ensuremath{\surd}}}1%
        {≈}{{\ensuremath{\approx}}}1%
        {≠}{{\ensuremath{\neq}}}1%
        {✓}{{\ensuremath{\checkmark}}}1
    },
  },
}
\newtcblisting{modelinput}{
  modeltranscript,
  title={Model Input}
}
\newtcblisting{modelinputsystem}{
  modeltranscript,
  title={Model Input (System)}
}
\newtcblisting{modelinputuser}{
  modeltranscript,
  title={Model Input (User)}
}
\newtcblisting{modeloutput}{
  modeltranscript,
  title={Model Output}
}

\usepackage{etoolbox}

\makeatletter
\AfterEndEnvironment{definition}{\@afterindentfalse}
\AfterEndEnvironment{theorem}{\@afterindentfalse}
\AfterEndEnvironment{lemma}{\@afterindentfalse}
\AfterEndEnvironment{proposition}{\@afterindentfalse}
\AfterEndEnvironment{corollary}{\@afterindentfalse}
\makeatother

\title{Tracing the Scientific Discovery\\ Capacity of LLMs to Their Training Data}

\author{
    \small
    Arda Uzunoglu$^{1}$,
    Mahler Revsine$^{1}$,
    Michael Schatz$^{1}$,
    Benjamin Van Durme$^{1,2}$ \\[0.3em]
    \small
    Mehrdad Farajtabar$^{3}$,
    Daniel Khashabi$^{1}$,
    Eric Nalisnick$^{1}$ \\[1em]
    \small
    $^{1}$Johns Hopkins University
    \quad
    $^{2}$Microsoft
    \quad
    $^{3}$Apple \\[0.6em]
    \small
    \href{https://github.com/ardauzunoglu/discovery}{GitHub}
    \quad \quad
    \href{https://huggingface.co/collections/ardauzunoglu/discovery}{Hugging Face}
}

\date{}

\begin{document}
\maketitle
\vspace{1.1em}

\begin{adjustwidth}{0.35in}{0.35in}
\begin{center}
    \bfseries Abstract
\end{center}
\vspace{0.3em}

\noindent
Large language models (LLMs) show growing promise for scientific discovery, yet basic questions remain about what enables reliable discovery. \textbf{How can we trace the scientific discovery capacity of LLMs to their training data?} We study this question as one of \emph{missing probability mass} and develop a theoretical account in which monofacts---the facts observed once during training---set the discovery capacity. In materials science and regulatory genomics, we directly manipulate monofact rates in training sets and find that more monofacts consistently yield more discoveries across two open-weight model families, sampling strategies, and reasoning budgets. Further experiments show the relationship weakens under miscalibration, cannot be explained by training-set diversity, and can transfer across scientific domains when paired with in-domain data. Together, our results identify monofact rate in training data as a key factor governing discoveries in language models.
\end{adjustwidth}

\section{Introduction}
Isaac Newton revolutionized our understanding of the physical world through his laws of motion and universal gravitation. Yet he also devoted substantial effort to alchemy and searching for hidden prophecies in biblical texts--pursuits that we now regard as deeply misguided \citep{gleick2007isaac}. Newton's example illustrates a fundamental tension in scientific inquiry: the pursuit of novel ideas can lead to extraordinary breakthroughs, but also to profoundly mistaken conclusions. This tension is becoming increasingly relevant to AI-powered science as large language models (LLMs) begin proposing and evaluating novel hypotheses \citep{gottweis2025coscientist}, discovering new mathematical constructions \citep{romeraparedes2024mathematical}, and contributing to solutions of open scientific problems \citep{openai2026scientificcollaborator,openai2026unitdistance}.\footnote{The pace is striking: during our work, frontier models contributed to a counterexample to the Jacobian conjecture~\citep{alpoge2026jacobian} and to a finite-time blowup construction for 3D Navier--Stokes~\citep{openai2026navierstokes}.} Making such discoveries, however, requires LLMs to venture beyond what they have already observed---precisely the same statistical regime in which hallucinations arise. Yet training data provides only observations of what is already known. This raises a fundamental question: \emph{how can we trace the scientific discovery capacity of LLMs to their training data?}

We approach this question through the \emph{missing mass} framework of \citet{kalai2024calibrated}, which links training set properties to the model's penchant for generating novel statements. They show that the Good--Turing estimate of missing mass~\citep{good1953population}, given by the prevalence of \emph{monofacts} (i.e., statements observed exactly once), provides a lower bound on the hallucination rate. However, discoveries, by definition, must also be drawn from the missing mass.  Yet \citet{kalai2024calibrated}'s framework treats unseen statements as equally likely to be true, offering no basis for distinguishing discoveries from falsehoods. Scientific domains, in contrast, contain structure that makes some unseen candidates more plausible than others \citep{meredig2018extrapolation,merchant2023scaling}. We capture this structure through a \emph{frontier of possible truths}, extending the framework to characterize scientific discovery. Our analysis establishes a \emph{discovery--hallucination exchange rate} governed by frontier ambiguity (Proposition~\ref{prop:h_d_exchange}) and, under calibration assumptions, establishes both upper and lower bounds on \emph{discovery} in terms of the same monofact statistic (Theorems~\ref{thm:discov_capac} and~\ref{thm:discov_floor}). Therefore, we extend the missing-mass perspective from hallucination to scientific discovery.

We empirically verify our framework in two realistic discovery tasks: materials science and regulatory genomics. For the former, we train LLMs to generate chemical formulas satisfying specified material properties. For the latter, we train LLMs to propose DNA sequence modifications that achieve target regulatory activity profiles. In both tasks, we measure discovery through the exact recovery of target outputs from test sets that postdate the models' knowledge cutoffs, requiring models to make predictions for samples they could not have seen.  To test the central claims of Theorems \ref{thm:discov_capac} and \ref{thm:discov_floor}---that monofacts drive discovery---we directly manipulate monofact rate by training models on datasets with controlled monofact rates (\S\ref{subsec:setup}). Across two model families, training stages, and evaluation settings, higher monofact rate consistently yields more discoveries (\S\ref{subsec:results}). Further experiments show that this relationship persists across reasoning budgets (\S\ref{subsec:effort-ablation}), weakens when repeated training increases calibration error (\S\ref{subsec:calibration-error}), cannot be explained by training-set diversity (\S\ref{subsec:or}), and transfers across scientific domains when paired with in-domain data (\S\ref{subsec:cross-domain}). Together, our results identify monofact rate as a key factor governing an LLM's discovery capacity.

\section{Background: Hallucination Bounds from Monofact Rates}
\label{sec:background}
We begin by summarizing the framework of \citet{kalai2024calibrated}, who formulate hallucination as missing mass and connect it to the Good--Turing estimator \citep{good1953population}. Their key result is that a calibrated LLM should generate false statements \emph{at least} as often as novel facts appear in the training set. This is an important observation since (i) hallucination rates can be estimated from simple statistics of the training data, provided the LLM is sufficiently well-calibrated, and (ii) hallucinations are driven by the long tail of the training distribution.  We provide the formal details below and offer a broader discussion of related work on long tail distributions, hallucination, and LLM-based scientific discovery in App.~\ref{sec:prior-work}. 

We first define calibration of an LLM distribution with respect to a distribution over statements. Let $\mathcal{S}$ be a finite space of semantic statements.  \citet{kalai2024calibrated} consider these statements to be at the level of semantics, meaning that paraphrases of the same statement do not have different representations in $\mathcal{S}$ \citep{kuhn2023semantic,nakkiran2026trained}.  Let $f \in \Delta(\mathcal{S})$ be the distribution induced by an LLM under unconditional generation, where $\Delta(\mathcal{S})$ denotes all possible distributions over $\mathcal{S}$. Let $\mathcal{W}$ be a \emph{world prior}, a distribution over document distributions, let $\mathfrak{D} \sim \mathcal{W}$ be a distribution over documents, and let $p_{\mathfrak{D}} \in \Delta(\mathcal{S})$ be the induced distribution over statements, which places probability mass only on \emph{true} statements.  The set of true statements is $\mathcal{T} \triangleq \{\rvs : p_{\mathfrak{D}}(\rvs) > 0\}$, and the set of hallucinations is its complement $\mathcal{H} \triangleq \mathcal{S} \setminus \mathcal{T}$, which captures all false statements.  \citet{kalai2024calibrated} define generative calibration with respect to $p_{\mathfrak{D}}$ and a partition of $\mathcal{S}$ as follows:

\begin{definition}[Generative Calibration] For $p_{\mathfrak{D}} \in \Delta(\mathcal{S})$ and a partition $\pi \in \Pi(\mathcal{S})$, a model $f$ is calibrated w.r.t.~$p_{\mathfrak{D}}$ if $f(\vs_0) = \mathbb{E}\left[ p_{\mathfrak{D}}(\rvs) \ | \ \rvs \in \pi(\vs_0) \right] =  \sum_{\rvs \in \pi(\vs_0) } p_{\mathfrak{D}}(\rvs) /|\pi(\vs_0)|$, for all $\vs_{0} \in \mathcal{S}$, where $\pi(\vs_0)$ denotes the partition to which $\vs_{0}$ is assigned and $|\pi(\vs_0)|$ is its cardinality. 
\end{definition} The granularity of $\pi$ determines calibration strength. If $\pi$ groups all statements together, even a constant $f$ over $\mathcal{S}$ can be calibrated, whereas increasingly granular partitions force $f$ to approach $p_{\mathfrak{D}}$ everywhere. In supervised learning, $\pi$ is often specified by the model's confidence in its top-ranked class \citep{guo2017calibration}, and \citet{kalai2024calibrated}'s results similarly require that the partition be a coarsening based on $f$, which we denote as $\pi_f$. 

Next, consider a set $\mathcal{D}$ of $N$ documents containing statements. A key statistic is $\mathcal{D}$'s \textit{monofact rate}:

\begin{definition}[Monofact Rate] Let $n_{\mathcal{D}}(\rvs)$ denote the number of occurrences of statement $\rvs$ in the training set $\mathcal{D}$. The monofact rate is then
$\hat{\mu} \triangleq
|\{\rvs \in \mathcal{S} : n_{\mathcal{D}}(\rvs) = 1\}| / N.$
\end{definition} 
\citet{kalai2024calibrated} assume that (i) there exists one statement per document, (ii) true statements have non-zero probability while false statements never occur ($\mathcal{D} \sim \mathfrak{D}$), (iii) $\mathcal{S}$ contains exponentially more false statements than true ones, (iv) conditioned on the training set $\mathcal{D}$, every unobserved statement is equally likely to be true and has equal expected probability under $p_{\mathfrak{D}}$ (i.e., regularity).  

Under these assumptions, \citet{kalai2024calibrated} derive the following finite-sample lower bound:

\begin{theorem}[KV-Bound] The  probability $f$ assigns to $\mathcal{H}$ is bounded, with probability $1-\delta$, as:
$$f(\mathcal{H}) \ \ge \ \hat{\mu} \ - \ \mathrm{CE}(f, p_{\mathfrak{D}}; \pi_f) \ - \ \frac{3 e^{-m}}{\delta} \ - \ \sqrt{\frac{6 \ln\{ 6/\delta \}}{N}},$$ where $\hat{\mu}$ is the Good-Turing monofact estimator, $\mathrm{CE}(f, p_{\mathfrak{D}}; \pi_f) \in [0, 1]$ denotes the calibration error of $f$, $m$ is a sparsity parameter from assumption (iii), $\delta$ is the bound's confidence, and $N$ is the number of training documents.
\end{theorem} 
See \citet{kalai2024calibrated} for the proof and a matching upper bound. The monofact rate estimates missing probability mass via Good--Turing \citep{good1953population}: more unique elements in a sample imply a greater chance of unseen elements in future samples. Assumptions (iii) and (iv) then imply that these future, novel samples are likely to lie in $\mathcal{H}$.

\citet{kalai2024calibrated} distinguish \textit{arbitrary} facts, which must be learned from observations, from \textit{systematic} facts, whose truth follows from underlying rules. The above theory applies to the former: a fact such as ``\textit{The Shining} was published in 1977'' must be learned through repetition, since there are no underlying rules that dictate that \textit{The Shining}'s publication date must be 1977. On the other hand, inequalities such as $2.345754 < 8.754$ can be generalized from structure even if never observed directly. Using a series of simulations, \citet{miao2025hallucination} empirically show that varying training set novelty drives hallucination as predicted by \citet{kalai2024calibrated}.

\section{Monofacts Determine the Scientific Discovery Capacity of LLMs}
\label{sec:discovery}

We now re-purpose the KV-Bound to characterize \emph{scientific discovery}. By definition, discoveries are absent from the training data and must therefore come from the \textit{missing mass} estimated by the Good--Turing statistic.  One interpretation of \citet{kalai2024calibrated} is that discoveries are simply systematic facts, and therefore fact counts do not play a role.  However, we believe discoveries do not cleanly fit into either category, as they are often made through \textit{abductive reasoning}---formulating hypotheses from plausible guesses---rather than straightforward deduction or induction. Or, as Isaac Newton put it, ``No great discovery was ever made without a bold guess.''  To go a step further, famous thought experiments such as Einstein's clock tower and Kekulé's self-cannibalizing snake were a subconscious fusion of scientific intuition and dream-like imagery.

To incorporate discovery into the KV framework, we first define the set of observed statements as $\mathcal{O} \triangleq \left\{ \ \rvs \ | \ \rvs \in \mathcal{D} \ \right\}$ and the unobserved statements as its complement $\mathcal{U} \triangleq \mathcal{S} \setminus \mathcal{O}$. A \emph{discovery} is then a statement from $\mathcal{T} \cap \mathcal{U}$.  The proof of the KV-Bound reveals that, under \citet{kalai2024calibrated}'s assumptions, discovery is impossible for \emph{any} learning algorithm.  The pivotal step bounds the model's expected mass on unobserved truths as $\mathbb{E}\left[ f(\mathcal{T} \cap \mathcal{U}) \ | \ \mathcal{D} \right] = \sum_{\rvs \in \mathcal{U}} f(\rvs) \ \mathbb{P}\left( \rvs \in \mathcal{T} \ | \ \mathcal{D} \right) \le \max_{\rvs \in \mathcal{U}} \mathbb{P}\left( \rvs \in \mathcal{T} \ | \ \mathcal{D} \right)$, and the regularity condition (iv), combined with sparsity (iii), caps this maximum at roughly $e^{-m}$, where $m$ quantifies the exponential sparsity of true statements. Thus, the original assumptions leave essentially no probability mass for discovery.

\paragraph{Discovery frontier.} Discovery must dwell in violations of this regularity, and thus we revise KV-theory to posit a three-band structure on the statement space.  This structure is motivated by scientific practice: discovery lives on a relatively small \emph{frontier} of conjectures, beyond which lies a vast expanse of falsehoods.  

\begin{definition}[Frontier Structure]  Conditioned on the training set $\mathcal{D}$, the statement space decomposes as $\mathcal{S} = \mathcal{O} \ \cup \ \mathcal{C} \ \cup \ \mathcal{R}$, where $\mathcal{O}$ is the observed band, $\mathcal{C} \subseteq \mathcal{U}$ is the \emph{candidate band} (the frontier), and $\mathcal{R} \triangleq \mathcal{U} \setminus \mathcal{C}$ is the arbitrary band.  For constants $0 < \alpha \le \beta < 1$, the world prior $\mathcal{W}$ is $(\alpha, \beta)$-separated if for all $\mathcal{D}$: $\alpha \ \le \ \mathbb{P}\left( \rvs \in \mathcal{T} \ | \ \mathcal{D} \right) \ \le \ \beta, \text{ for all } \rvs \in \mathcal{C}. $  The arbitrary band $\mathcal{R}$ satisfies the original regularity condition (iv) of \citet{kalai2024calibrated}, with sparsity parameter $m$ applying to $\mathcal{R}$ alone. 
\end{definition} 
Statements with posterior probability near $1$ are \emph{obvious}: their truth is already determined by the training data and are captured by $\mathcal{O}$'s closure.  Statements whose posterior is near $0$ are indistinguishable from the arbitrary corruptions populating $\mathcal{R}$.  Discovery lives in the intermediate regime, $\mathcal{C}$, where the data elevates a candidate above noise without fully settling its truth.  Our $[\alpha, \beta]$-frontier is similar to the \textit{Zone of Proximal Development} \citep{shabani2010vygotsky} from educational psychology, which concerns problems between what a student can and cannot solve. However, the frontier does not fully separate discoveries from hallucinations, as the following result makes clear.

\begin{proposition}[Discovery--Hallucination Exchange Rate]\label{prop:h_d_exchange}
Assume an $(\alpha, \beta)$-separated world prior and assumptions (i)--(iii) of \citet{kalai2024calibrated}.  For any learning algorithm producing $f$ with $f(\mathcal{C}) > 0$, define the \emph{realized frontier posterior} to be
$$ \gamma_f \ \triangleq \ \frac{ \mathbb{E}\left[ f(\mathcal{T} \cap \mathcal{C}) \ | \ \mathcal{D} \right] }{ f(\mathcal{C}) } \ = \ \sum_{\rvs \in \mathcal{C}} \frac{f(\rvs)}{f(\mathcal{C})} \ \mathbb{P}\left( \rvs \in \mathcal{T} \ | \ \mathcal{D} \right). $$
The ratio of expected discovery mass to expected frontier hallucination mass is bounded by:
$$ \frac{\alpha}{1 - \alpha} \ \le \ \frac{ \mathbb{E}\left[ f(\mathcal{T} \cap \mathcal{C}) \ | \ \mathcal{D} \right] }{ \mathbb{E}\left[ f(\mathcal{H} \cap \mathcal{C}) \ | \ \mathcal{D} \right] } \ = \ \frac{\gamma_f}{1 - \gamma_f} \ \le \ \frac{\beta}{1 - \beta}. $$
\end{proposition}

\begin{proof}
Conditioned on $\mathcal{D}$, the model $f$ is fixed. So, by linearity of expectation $\mathbb{E}\left[ f(\mathcal{T} \cap \mathcal{C}) \ | \ \mathcal{D} \right] = \sum_{\rvs \in \mathcal{C}} f(\rvs) \, \mathbb{P}\left( \rvs \in \mathcal{T} \ | \ \mathcal{D} \right) = \gamma_f \, f(\mathcal{C})$.  Since $\mathcal{T}$ and $\mathcal{H}$ partition $\mathcal{C}$, $\mathbb{E}\left[ f(\mathcal{H} \cap \mathcal{C}) \ | \ \mathcal{D} \right] = f(\mathcal{C}) - \gamma_f \, f(\mathcal{C})$.  As $\gamma_f$ is a convex combination of the posteriors $\mathbb{P}\left( \rvs \in \mathcal{T} \ | \ \mathcal{D} \right)$ over $\rvs \in \mathcal{C}$, the band gives $\gamma_f \in [\alpha, \beta]$.  Dividing the two identities yields the ratio $\gamma_f / (1 - \gamma_f)$, and the outer bounds follow since $t \mapsto t/(1-t)$ is increasing on $[0, 1)$.
\end{proof}

This shows that models that discover must also \emph{hallucinate at a proportional rate}. A model with frontier posteriors $\alpha = \beta = 1/2$ can expect one discovery per hallucination.  Shifting $\beta \to 1$ permits a more favorable ratio but skews the frontier towards $\mathcal{O}$,~i.e. incremental science. On the other hand, shifting $\alpha \to 0$ skews it toward $\mathcal{R}$, i.e. ambitious science with a small signal-to-noise ratio.  

\paragraph{Monofacts set discovery capacity.} With this frontier structure in place, we next show how the monofact rate can determine a model's capacity for discovery.  Call a partition $\pi_f$ \emph{frontier-separated} if every cell containing a statement of $\mathcal{C}$ contains no observed statement.  This prevents observed probability mass from entering frontier cells under coarsening.  We then obtain the following upper bound on the expected probability the LLM places on discoveries:

\begin{theorem}[Discovery Capacity]\label{thm:discov_capac}
Assume (i)--(iii) of \citet{kalai2024calibrated}, an $(\alpha, \beta)$-separated world prior, and that $\pi_f$ is frontier-separated.  With probability $1 - \delta$ over the draw of the training set, the following bound holds for any model $f$ on the expected probability of discovery:
$$ \mathbb{E}\left[ f(\mathcal{T} \cap \mathcal{C}) \ | \ \mathcal{D} \right] \ \le \ \beta \left( \ \hat{\mu} \ + \ \mathrm{CE}(f, p_{\mathfrak{D}}; \pi_f) \ + \ \sqrt{\frac{3 \ln\{ 3/\delta \}}{N}} \ \right). $$
\end{theorem}

\begin{proof} The $[\alpha, \beta]$-band gives $\mathbb{E}\left[ f(\mathcal{T} \cap \mathcal{C}) \ | \ \mathcal{D} \right] \le \beta f(\mathcal{C})$ by the termwise argument of Proposition \ref{prop:h_d_exchange}.  The calibration error bounds the model's deviation from the coarsening, $f(\mathcal{C}) \le p_{\mathfrak{D}}^{\pi_f}(\mathcal{C}) + \mathrm{CE}(f, p_{\mathfrak{D}}; \pi_f)$, and frontier separation ensures the cells meeting $\mathcal{C}$ lie entirely in $\mathcal{U}$, and hence $p_{\mathfrak{D}}^{\pi_f}(\mathcal{C}) \le p_{\mathfrak{D}}(\mathcal{U})$, the missing mass.  The Good--Turing concentration underlying the KV-Bound bounds the missing mass by $\hat{\mu}$ plus the stated deviation term with probability $1-\delta$. 
\end{proof}

\paragraph{Monofacts drive discovery.} Notably, Theorem~\ref{thm:discov_capac} is an upper bound, and therefore characterizes discovery capacity rather than guaranteed discovery.  A lower bound would instead quantify what a model \emph{will} discover (in expectation), leading to a discovery floor.  Deriving this bound requires another parameter, which captures the rate at which the model samples from $\mathcal{C}$ vs $\mathcal{R}$.  Define the model's \emph{frontier leakage} as the fraction of its unobserved mass that leaks into $\mathcal{R}$ and away from $\mathcal{C}$, which is important since discoveries are concentrated in $\mathcal C$ rather than in $\mathcal R$:
$$ \rho_f \ \triangleq \ \frac{ f(\mathcal{R}) }{ f(\mathcal{U}) } \ \in \ [0, 1], \qquad f(\mathcal{C}) \ = \ (1 - \rho_f) \, f(\mathcal{U}). $$  Thus a \emph{useful} model that has the ability to generalize to $\mathcal{C}$ should have a small value for $\rho_f$.  Importantly, $\rho_f$ may be estimated by repeatedly sampling from the model. Since generations from the diffuse arbitrary band recur less often than frontier generations, Good--Turing estimation on model samples offers a way to estimate $\rho_f$, which we apply in App.~\ref{app:b4}, building on recent work using missing-mass estimators for uncertainty quantification in black-box generative models~\citep{noorani2025conformal}. Using $\rho_f$, we obtain the following lower bound.

\begin{theorem}[Discovery Floor]\label{thm:discov_floor}
Assume (i)--(iii) of \citet{kalai2024calibrated}, an $(\alpha, \beta)$-separated world prior, and that $\pi_f$ is frontier-separated.  With probability $1 - \delta$ over $\mathfrak{D} \sim \mathcal{W}$ and $\mathcal{D} \sim \mathfrak{D}$, the following bound holds for any model $f$ on the expected probability of discovery:
$$ \mathbb{E}\left[ f(\mathcal{T} \cap \mathcal{C}) \ | \ \mathcal{D} \right] \ \ge \ \alpha \, (1 - \rho_f) \left( \ \hat{\mu} \ - \ \mathrm{CE}(f, p_{\mathfrak{D}}; \pi_f) \ - \ \frac{3 e^{-m}}{\delta} \ - \ \sqrt{\frac{6 \ln\{ 6/\delta \}}{N}} \ \right). $$
\end{theorem}

\begin{proof}
Proposition~\ref{prop:h_d_exchange} gives $\mathbb{E}\left[ f(\mathcal{T} \cap \mathcal{C}) \ | \ \mathcal{D} \right] = \gamma_f \, f(\mathcal{C}) \ge \alpha \, f(\mathcal{C}) = \alpha (1 - \rho_f) f(\mathcal{U})$.  Therefore it suffices to lower bound $f(\mathcal{U})$, which we do by following the proof of the KV-Bound up to the point at which it subtracts the unobserved-truth mass.  Write $q \triangleq p_{\mathfrak{D}}^{\pi_f}$ for the coarsening.  Since $\mathrm{CE}(f, p_{\mathfrak{D}}; \pi_f) = \| q - f \|_{\mathrm{TV}}$, we have $f(\mathcal{U}) \ge q(\mathcal{U}) - \mathrm{CE}(f, p_{\mathfrak{D}}; \pi_f)$.  Next, $q(\mathcal{U}) = p_{\mathfrak{D}}(\mathcal{U}) - (p_{\mathfrak{D}}(\mathcal{U}) - q(\mathcal{U}))$, and coarsening can move true mass off $\mathcal{U}$ only within cells that also contain observed statements. Lemma~2 of \citet{kalai2024calibrated} gives the pointwise bound $(p_{\mathfrak{D}}(\mathcal{U}) - q(\mathcal{U}))_+ \le \sum_{B \cap \mathcal{O} \ne \emptyset} |B \cap \mathcal{O}| \, p_{\mathfrak{D}}(B \cap \mathcal{U}) / |B \cap \mathcal{U}|$.  Under frontier separation, no such cell meets $\mathcal{C}$, so $B \cap \mathcal{U} \subseteq \mathcal{R}$ in every term, and regularity (iv) on $\mathcal{R}$ makes each ratio have conditional expectation at most $\mathbb{E}[p_{\mathfrak{D}}(\mathcal{R}) \ | \ \mathcal{D}] / |\mathcal{R}| \le 1 / |\mathcal{R}|$.  Since $\mathcal{O} \subseteq \mathcal{T}$ by (ii) and $|\mathcal{T}| \le e^{-m} |\mathcal{R}|$ by (iii) applied to $\mathcal{R}$, the conditional expectation of the sum is at most $|\mathcal{O}| / |\mathcal{R}| \le e^{-m}$, and Markov's inequality gives $p_{\mathfrak{D}}(\mathcal{U}) - q(\mathcal{U}) \le 3 e^{-m} / \delta$ with probability $1 - \delta/3$ over the posterior draw of $p_{\mathfrak{D}}$.  Finally, the Good--Turing lower tail (Corollary~6 of \citet{kalai2024calibrated}) gives $p_{\mathfrak{D}}(\mathcal{U}) \ge \hat{\mu} - \sqrt{6 \ln\{6/\delta\} / N}$ with probability $1 - \delta/3$ over the training draw, leading to a union bound of the two events.
\end{proof}
As $\rho_f$ approaches $1$, the discovery floor vanishes, recovering the regime of \citet{kalai2024calibrated} in which unobserved probability mass is predominantly allocated to hallucinations. Together, Theorems~\ref{thm:discov_capac} and~\ref{thm:discov_floor} bound expected discovery in terms of the empirical monofact rate $\hat{\mu}$, with a ceiling-to-floor ratio of $\beta/(\alpha(1-\rho_f))$ when calibration and finite-sample terms are negligible. The frontier parameters $\alpha$ and $\beta$ govern the discovery--hallucination exchange rate, while $\rho_f$ captures the fraction of unobserved probability mass allocated outside the discovery frontier. 

\begin{figure}
    \centering
    \includegraphics[width=0.9\linewidth]{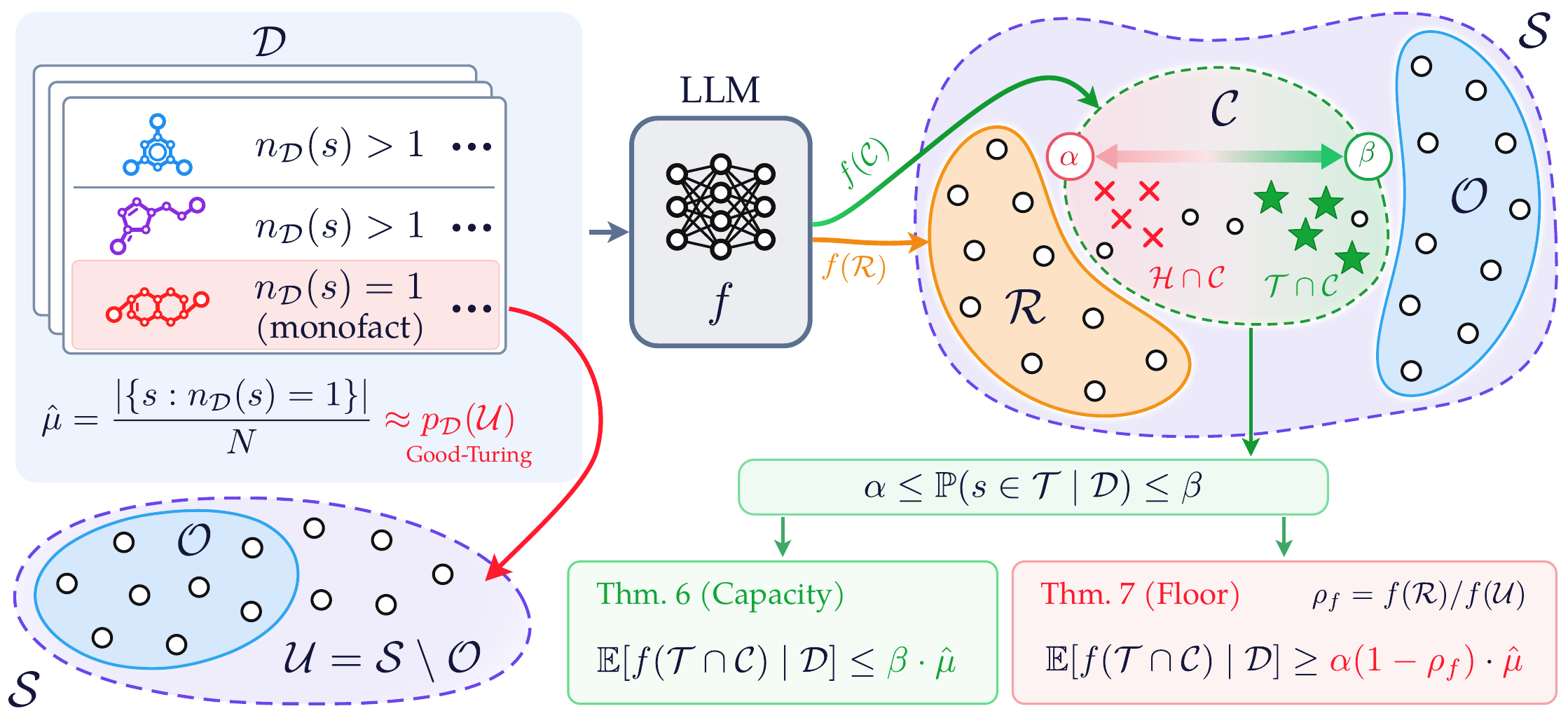}
    \caption{\textbf{Monofacts and discovery capacity.} The monofact rate in the training data estimates the missing probability mass via Good--Turing. A language model allocates probability to unseen statements, including the candidate frontier $\mathcal{C}$ and the arbitrary region $\mathcal{R}$. Within the frontier, discoveries and hallucinations coexist, with their relative prevalence governed by the parameters $\alpha$ and $\beta$.
    }
    \label{fig:monofacts_frontier}
\end{figure}

Overall, our theoretical framework inverts the intuition from \citet{kalai2024calibrated} that monofacts are undesirable because they lead to hallucination.  We show that \emph{monofacts also allow LLMs to draw from the missing mass in which discoveries reside}.  Figure~\ref{fig:monofacts_frontier} summarizes our contributions, connecting the monofact rate to missing mass, the discovery frontier, and the resulting bounds on expected discovery.

\section{Experiments}
\label{sec:exp}
We empirically test the central consequence of Theorems \ref{thm:discov_capac} and \ref{thm:discov_floor}, that monofacts drive discovery, in the domains of materials science and regulatory genomics.

\subsection{Setup: Tasks, Models, and Evaluation}
\label{subsec:setup}
\paragraph{Materials discovery.}
We first consider materials discovery, where the model receives a chemical system, target properties, and a sanitized composition-blind view of the reference crystal structure (App.~\ref{app:training-examples}), and must propose the exact material. To ensure that test-set successes correspond to genuine discoveries, we enforce a temporal train--test split using two snapshots of the \textit{Materials Project} \citep{jain2013materials}, taking the November 14, 2024 snapshot for training and formulas first appearing in the September 25, 2025 snapshot for evaluation.  We define rarity at the level of \emph{chemical systems}, where monofacts are chemical systems that appear exactly once in the training set.\footnote{A chemical system is the set of constituent elements of a material. A single chemical system can contain multiple materials with different structures and properties; for example, H$_2$O and H$_2$O$_2$ both belong to H--O.} We construct training sets with monofact rates of $r \in \{0.0,0.1,\ldots,1.0\}$, where a fraction $r$ of the training samples are monofacts. For each rate, we use a two-stage training pipeline.  For the first stage, we independently sample three datasets of 10K examples, fine-tuning the model for direct formula prediction.  For the second stage, the model continues training from the first stage checkpoints on a separate dataset of 4K examples with the same monofact rate $r$, using reasoning traces distilled from Claude Opus 5~\citep{opus5} (App.~\ref{app:opus5-distill}). This second stage more closely reflects practical LLM-assisted discovery workflows that involve explicit reasoning prior to hypothesis generation. Within each stage, all monofact rates use the same number of examples and are trained for one epoch.

\paragraph{Regulatory genomics discovery.} We next consider regulatory genomics, where the model predicts a single edit (a substitution or insertion, its position, and the new bases) to a 200-nucleotide reference DNA sequence that matches a specified regulatory activity across K562, HepG2, and SK-N-SH cells. We use the MPRA dataset of \citet{gosai2024machine}, released in October 2024. Because the dataset postdates the knowledge cutoffs of the models we use (see below), we randomly hold out examples for evaluation rather than constructing a temporal split. Unlike materials discovery, regulatory sequences do not admit a natural discrete grouping analogous to a chemical system. We thus define rarity using sequence similarity, where a sequence is a monofact if no other example in the training set overlaps with it on more than 150 nucleobases. We sweep $r \in \{0.0,0.1,\ldots,1.0\}$ and use the same two-stage training procedure, with three independently sampled 20K example datasets in the first stage and a separate 4K example dataset in the second.

\paragraph{Models.}
We use \textbf{\texttt{gpt-oss-20b}}~\citep{openai2025gptoss120bgptoss20bmodel} and \textbf{\texttt{gemma-3-12b-it}}~\citep{gemmateam2025gemma3technicalreport}, representing two open-weight model families with knowledge cutoffs in June 2024 and August 2024, respectively. Both cutoffs therefore precede the evaluation sets in our domains, supporting our notion of discovery.

\paragraph{Evaluation.}
Following the standard offline test of discovery \citep{sourati2023accelerating, yang2025moosechem, riebesell2025framework, tshitoyan2019unsupervised}, we operationalize discovery as exact recovery of the ground-truth material or DNA edit. We evaluate each model under deterministic decoding (\textbf{\texttt{pass@1}}) and under stochastic decoding with 16 generations per example (\textbf{\texttt{pass@16}}), counting success if any of the generations matches the target. For stochastic decoding, we use temperature $1.0$, top-$p=1.0$, no top-$k$ truncation, and a maximum generation length of 8192 tokens. We use medium reasoning effort throughout.

\paragraph{From theory to experiments.} Across our runs, $N$ and $\delta$ are fixed by design, while $\alpha$, $\beta$, and $m$ are fixed by the domain. In materials discovery, calibration error and $\rho_f$ also remain stable across monofact rates (Apps.~\ref{app:calibration} and~\ref{app:b4}). The monofact rate $\hat{\mu}$ is therefore the only term in our bounds that varies, so we attribute the observed trends to its increase.

\subsection{Finding: Discovery Increases with Higher Monofact Rate}
\label{subsec:results}
\begin{figure}[t]
    \centering

    \begin{subfigure}{\linewidth}
        \centering
        \includegraphics[width=\linewidth]{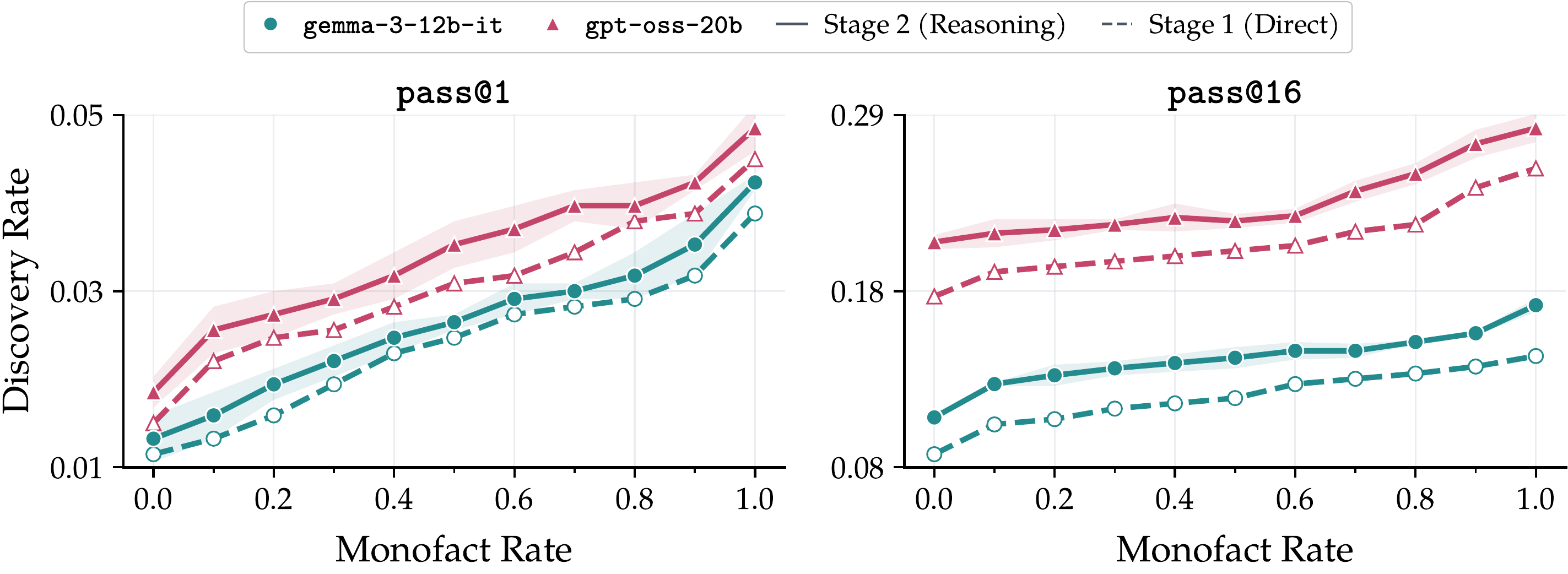}
        \caption{Materials discovery increases monotonically with the training monofact rate across all settings.}
        \label{fig:matsci}
    \end{subfigure}
    \begin{subfigure}{\linewidth}
        \centering
        \includegraphics[width=\linewidth]{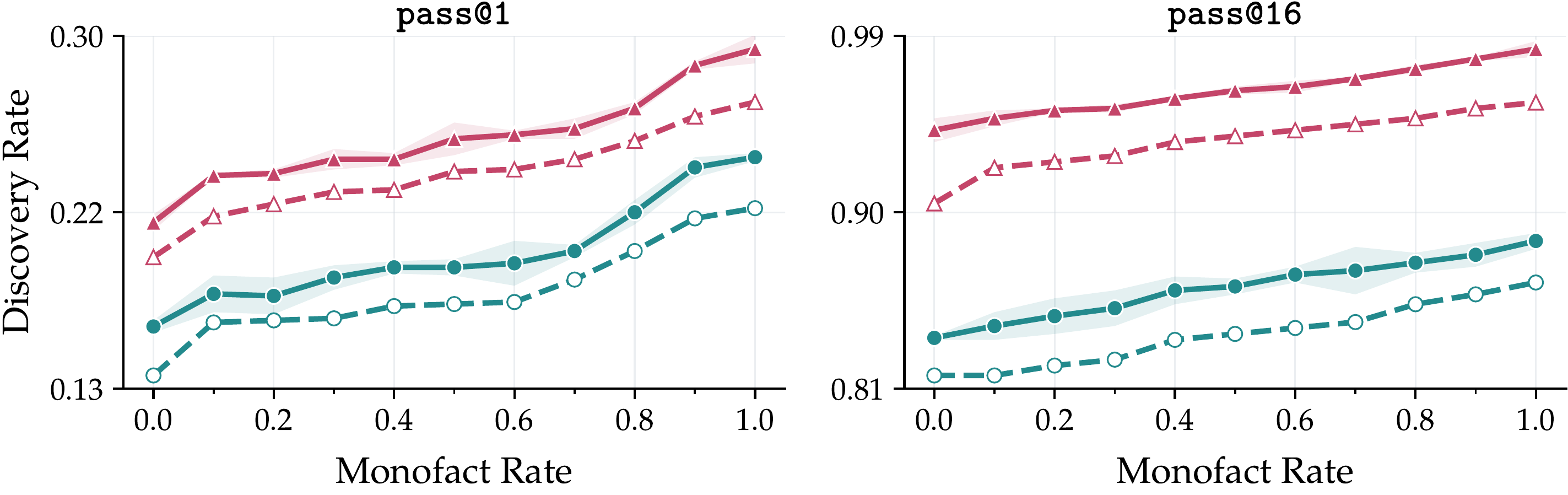}
        \caption{Regulatory genomics discovery increases monotonically with the training monofact rate across all settings.}
        \label{fig:bio}
    \end{subfigure}
    \caption{
    \textbf{Discovery increases monotonically with monofact rate, validating the predictions of Theorems~\ref{thm:discov_capac} and~\ref{thm:discov_floor}.} The relationship holds in \textbf{(a)} materials discovery and \textbf{(b)} regulatory genomics discovery across model families, training stages, and evaluation settings. Shaded regions denote standard error.
    }
    \label{fig:bio-matsci}
\end{figure}
\autoref{fig:bio-matsci} shows that discovery increases monotonically with the training monofact rate. This relationship holds in both scientific domains, for both model families, after both training stages, and under both \texttt{pass@1} and \texttt{pass@16} evaluation. Thus, the relationship between monofact rate and discovery is consistent across substantially different domains, models, and evaluation settings.

Notably, the monotonic relationship persists as absolute discovery rates increase toward saturation. For example, sampling 16 candidates substantially increases discovery rates compared to a single sample. Similarly, second-stage reasoning supervision improves upon first-stage performance. Nevertheless, neither changes the qualitative trends.

These results therefore support the discovery capacity prediction of Theorems~\ref{thm:discov_capac} and \ref{thm:discov_floor}, which bound discovery in terms of the monofact rate and permits a larger discovery budget as it increases. Since calibration error enters the discovery bounds, we additionally assess calibration of these models in App.~\ref{app:calibration}.

\subsection{Finding: Discovery is Coupled with Hallucination In-Distribution}
\label{subsec:increased-hal}

Discovery and hallucination both draw from the unobserved mass $\mathcal{U}$ (\S\ref{sec:discovery}). Our theory predicts that increasing the monofact rate expands this missing-mass budget (Theorems~\ref{thm:discov_capac} and \ref{thm:discov_floor}), while the frontier structure prevents the additional mass from being allocated exclusively to discoveries (Proposition~\ref{prop:h_d_exchange}). We therefore expect that, when training and evaluation distributions coincide, higher monofact rate increases both discovery and hallucination.

\begin{figure}[H]
    \centering
    \includegraphics[width=\linewidth]{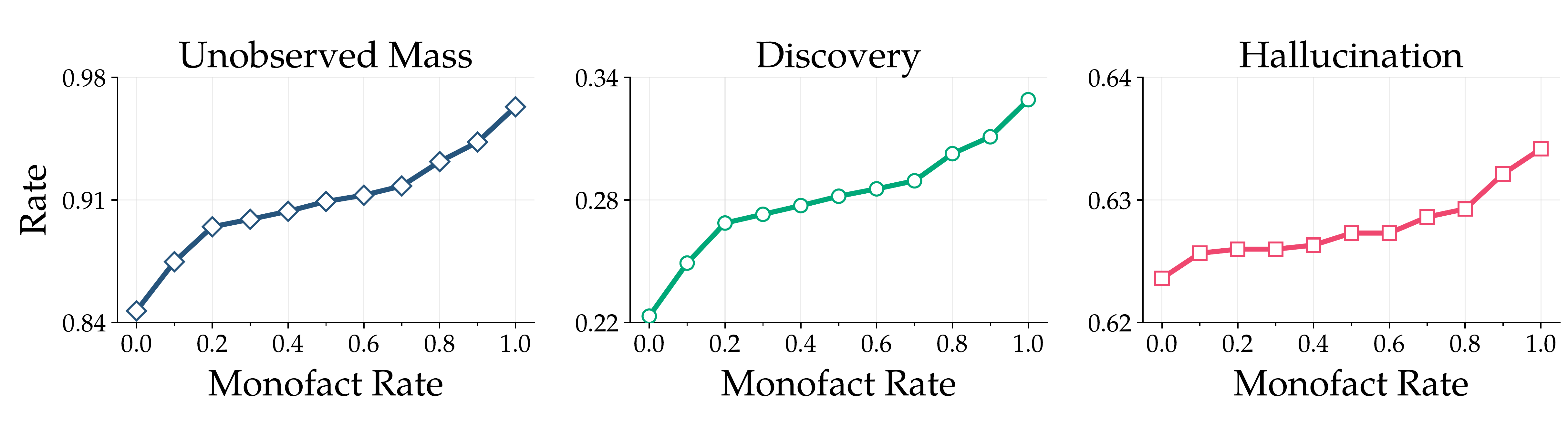}
    \caption{\textbf{Higher monofact rate increases both discovery and hallucination in-distribution (\texttt{pass@16}).} Unobserved mass is measured as the sum of discovery and hallucination.}
    \label{fig:inc-hall-unseen}
\end{figure}

We test this prediction in the materials discovery domain using \textbf{\texttt{gpt-oss-20b}}. We focus on \emph{Train (Unseen)}, which consists of held-out examples drawn from the same November 2024 Materials Project snapshot as the training data (\S\ref{subsec:setup}). This setting isolates in-distribution novelty, since the targets were not observed during training but come from the same underlying data distribution. \autoref{fig:inc-hall-unseen} shows that the model generates more unobserved mass as the training monofact rate increases. Both discovery and hallucination increase with this unobserved generation, consistent with the missing-mass view that a larger discovery budget also creates more opportunity for hallucination.

We defer the full decomposition to App.~\ref{app:missing-mass-decomp}, where we additionally analyze \emph{Train (Seen)} and the temporally shifted \emph{Test} sets and report the corresponding \texttt{pass@1} results.

\section{Characterizing Discovery Capacity}
\label{sec:add-exp}

We now evaluate the robustness of our findings across inference-time reasoning budgets, investigate the role of calibration error predicted by our theory, rule out training-set diversity as an alternative explanation, and test whether the monofact--discovery trend generalizes across scientific domains.

\subsection{Discovery Capacity Persists Across Reasoning Budgets}
\label{subsec:effort-ablation}

\paragraph{Motivation and setup.} Our experiments evaluate \texttt{gpt-oss-20b} at medium reasoning effort (\S\ref{subsec:setup}). We now test whether the observed relationship between monofact rate and discovery is simply a consequence of limited inference-time reasoning, namely, whether allocating more reasoning can compensate for the lower discovery rates at small monofact rates. Accordingly, we re-evaluate the second-stage checkpoints at low, medium, and high reasoning effort without retraining. 

\begin{figure}[H]
    \centering
    \includegraphics[width=\linewidth]{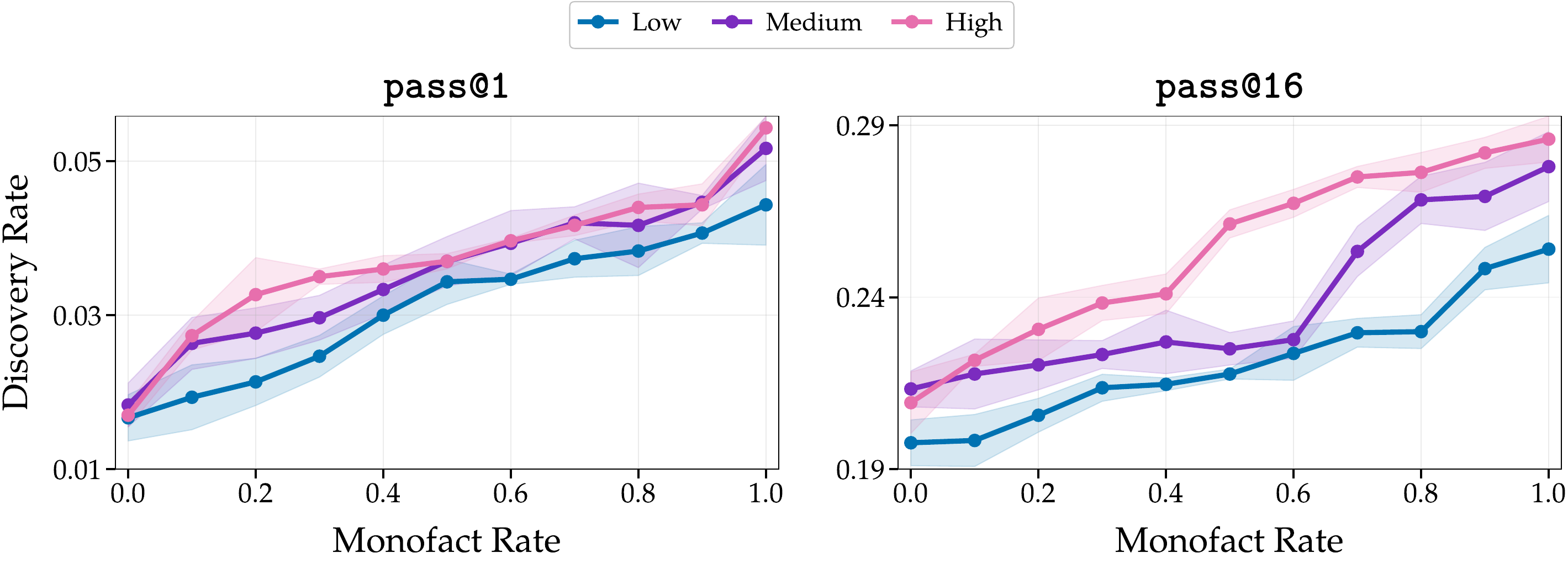}
    \caption{\textbf{Monofact--discovery relationship persists across different reasoning efforts.} Shaded regions denote standard error.}
    \label{fig:effort}
\end{figure}

\paragraph{Results.} \autoref{fig:effort} shows that the central trend is unchanged across different reasoning effort settings. For both \texttt{pass@1} and \texttt{pass@16}, discovery increases with the training monofact rate under all three settings. Notably, higher reasoning effort increases absolute discovery rates but does not alter the qualitative trend we observe. Thus, the relationship between monofact rate and discovery is robust to the inference-time reasoning budget.

\subsection{Miscalibration Weakens the Monofact--Discovery Relationship}
\label{subsec:calibration-error}

\paragraph{Motivation.} Theorems~\ref{thm:discov_capac} and~\ref{thm:discov_floor} bound discovery in terms of the monofact rate, but both bounds also depend on the calibration error. Since the discovery floor subtracts the calibration error from the monofact rate, our theory predicts that the monofact--discovery relationship should weaken when calibration error grows alongside the monofact rate. In our main experiments, however, calibration error remains nearly constant across monofact rates (App.~\ref{app:calibration}), leaving this prediction untested. At the same time, repeated training is known to shift language models from generalization toward memorization~\citep{xue2023repeat,aitchison2024overfit,miao2025hallucination}. We therefore test this prediction by repeating the unchanged training sets, which keeps $\hat{\mu}$ fixed but increases calibration error on the test set (App.~\ref{app:calibration}).

\paragraph{Setup.} Our main experiments train each model for one epoch (\S\ref{subsec:setup}). We now instead train \texttt{gpt-oss-20b} on materials discovery for three and five epochs in the second stage, initializing from one-epoch first-stage checkpoints with matching monofact rates.

\begin{figure}[H]
    \centering
    \includegraphics[width=\linewidth]{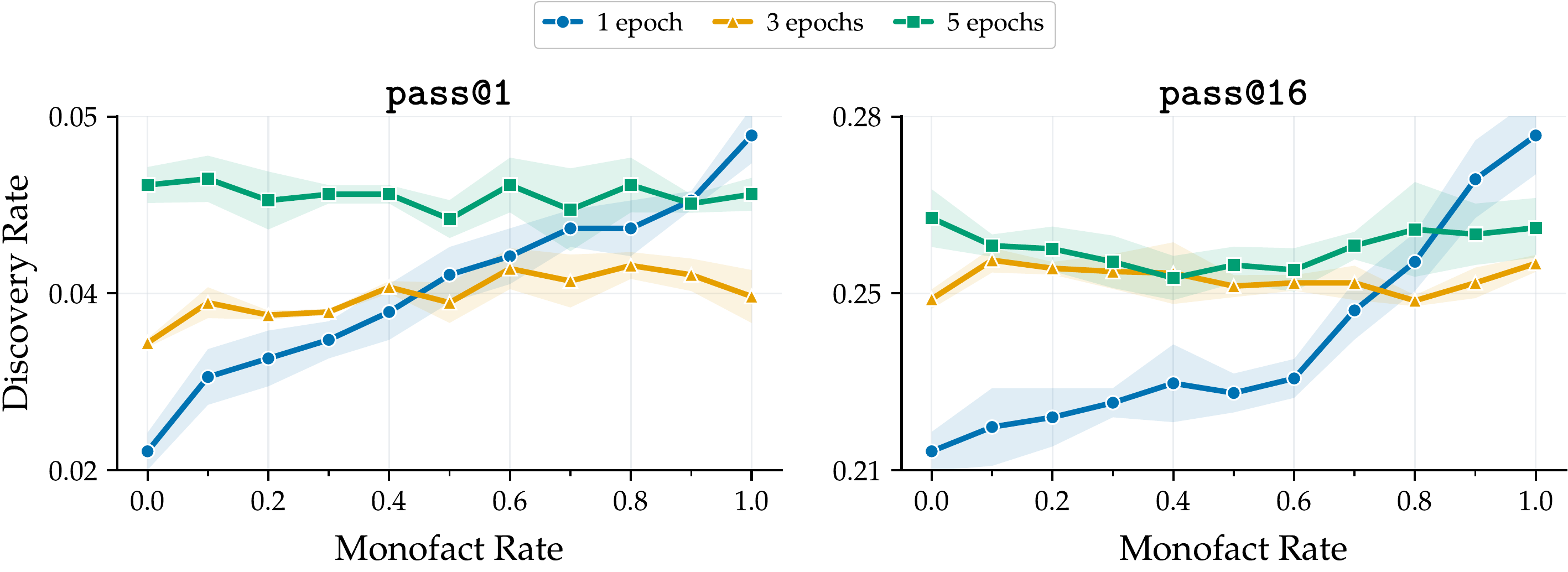}
    \caption{\textbf{Repeated training weakens the monofact--discovery relationship despite an unchanged monofact rate.} The x-axis denotes the monofact rate in the original training set, prior to repetition. Shaded regions denote standard error.}
    \label{fig:multiep}
\end{figure}

\paragraph{Results.} \autoref{fig:multiep}, along with App.~\ref{app:calibration}, supports this prediction. Under one-epoch training, ECE remains nearly flat and discovery increases with the monofact rate. Under three and five epochs, ECE instead increases with the monofact rate, and the monotonic relationship disappears, with discovery remaining roughly flat under both \textbf{\texttt{pass@1}} and \textbf{\texttt{pass@16}}. Additional epochs improve discovery at low monofact rates, but one-epoch training performs best at high monofact rates, where calibration error increases the most. Thus, once calibration error grows with the monofact rate, additional monofacts no longer translate into additional discoveries.

\subsection{Diversity Does Not Explain Monofact-Driven Discovery Capacity}
\label{subsec:or}

\begin{wraptable}[8]{r}{0.5\textwidth}
  \vspace{-0.75\baselineskip}
  \centering\footnotesize\setlength{\tabcolsep}{4pt}
  \begin{tabular}{lcc}
    \toprule
    Dataset & Monofact Rate & Diversity \\
    \midrule
    (Fe--O, Fe--O, Fe--O, Fe--O) & 0\%  & 25\% \\
    (Fe--O, Fe--O, Si--O, Si--O) & 0\%  & 50\% \\
    (Fe--O, Si--O, Si--O, Si--O) & 25\% & 50\% \\
    \bottomrule
  \end{tabular}
  \caption{\textbf{Monofacts vs.\ diversity.} Each row contains four training examples.}
\label{tab:monofact-diversity}
\end{wraptable}

\paragraph{Motivation.} A simpler explanation for our results is that monofact rate may be a proxy for \emph{training-set diversity}. For a training set of $N$ examples, let $n_s$ denote the number of examples belonging to chemical system $s$. We define diversity as the fraction of distinct chemical systems, $d = |\{s : n_s > 0\}|/N$, and the monofact rate as the fraction of examples whose chemical system appears exactly once, $r = \sum_s \mathbb{I}[n_s = 1]/N$. Although related, the two quantities are distinct. As \autoref{tab:monofact-diversity} shows, training sets can have identical diversity but different monofact rates, and vice versa. We therefore build new training sets that vary $r$ directly while holding diversity, dataset size, and training budget fixed, with every example seen once.

\paragraph{Setup.} We construct training sets of $N = 10$K examples drawn from the same $K = 5$K chemical systems, which fixes diversity at $d = K/N = 0.5$. For a monofact rate $r$, $rN$ chemical systems contribute a single example each, and the remaining $(1-r)N$ examples are distributed evenly across the other $K - rN$ systems. Since each monofact constitutes its own chemical system, it must be true that $r < d$, and we therefore consider $r \in \{0.0, 0.1, \ldots, 0.4\}$. We train \texttt{gpt-oss-20b} for one epoch on three independently sampled training sets per monofact rate following the first-stage training setup, and evaluate on the same test set as in \S\ref{sec:exp}. Consequently, if diversity drives the observed relationship, discovery should remain unchanged across monofact rates because diversity is fixed.

\begin{figure}[H]
    \centering
    \includegraphics[width=\linewidth]{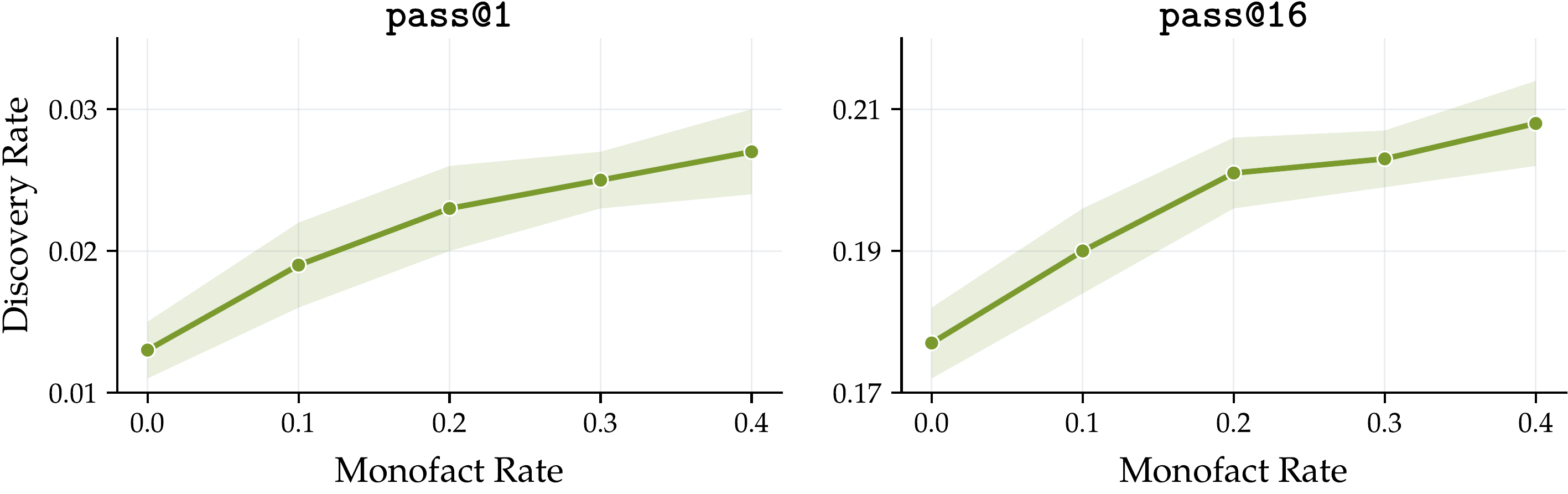}
    \caption{\textbf{Discovery increases with monofact rate at fixed diversity.} All training sets contain $N = 10$K examples from the same $K = 5$K chemical systems, fixing diversity at $d = 0.5$. Shaded regions denote standard error.}
    \label{fig:fixed-div}
\end{figure}

\paragraph{Results.} \autoref{fig:fixed-div} shows that discovery increases with the monofact rate despite fixed diversity, dataset size, and training budget. Thus, the observed gains reflect the effect of chemical systems that are observed only once, suggesting that training-set diversity alone does not account for the relationship between monofact rate and discovery.

\subsection{Discovery Capacity (Sometimes) Transfers Across Scientific Domains}
\label{subsec:cross-domain}

\paragraph{Motivation.} Our experiments so far train and evaluate within a single domain. In practice, however, in-domain discovery data is often scarce, while data from other scientific domains may be abundant. This raises the question of whether monofacts from one domain can expand discovery capacity in another, that is, whether monofact-driven discovery capacity is domain-specific or generalizable.

\paragraph{Setup.} We consider two settings, with in-domain materials data present (\emph{Mixed Domains}) and absent (\emph{Cross-Domain}). For \emph{Mixed Domains}, we repeat second-stage training with \texttt{gpt-oss-20b} using the 4K materials discovery dataset at monofact rate $1.0$ and 4K regulatory genomics datasets whose monofact rate varies from $0.0$ to $1.0$. For \emph{Cross-Domain}, we evaluate the regulatory genomics checkpoints from \S\ref{subsec:results} without additional training. We compare both against the second-stage checkpoint trained for materials discovery at monofact rate $1.0$ (\emph{In-Domain}).

\begin{figure}[H]
    \centering
    \includegraphics[width=\linewidth]{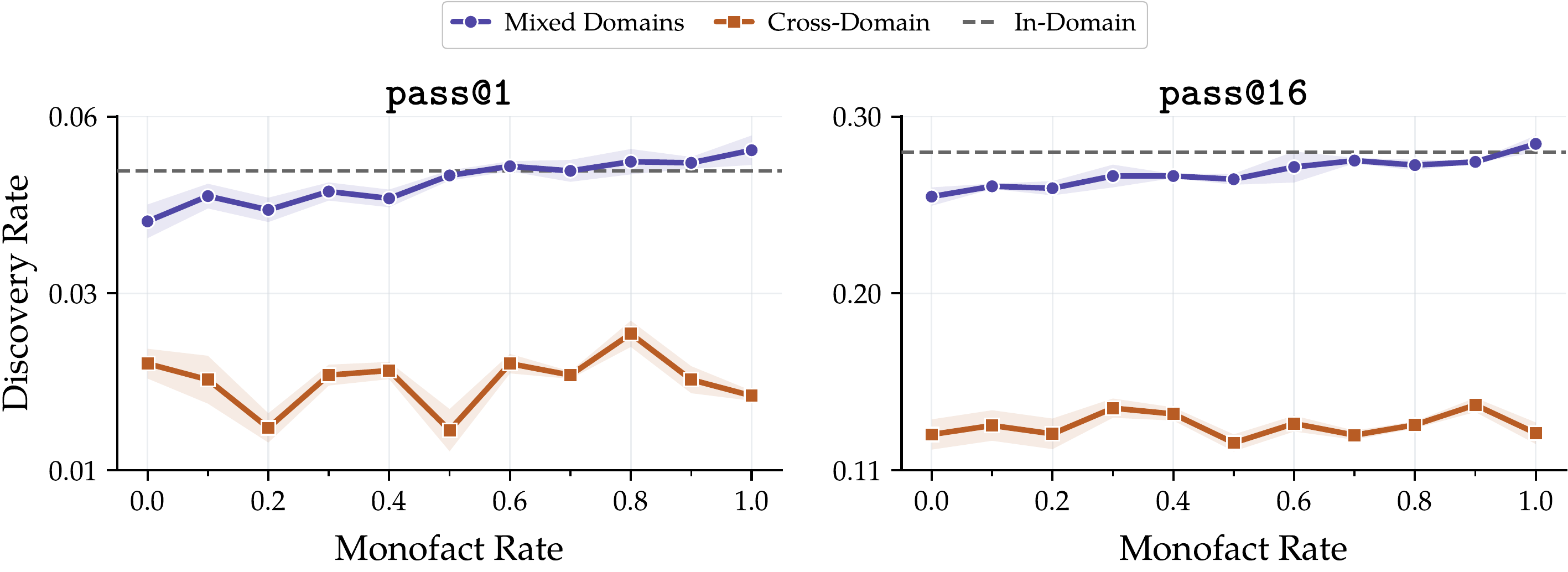}
    \caption{\textbf{Cross-domain monofact rate affects materials discovery when paired with in-domain training, but not on its own.} Shaded regions denote standard error.}
    \label{fig:cross-domain}
\end{figure}

\paragraph{Results.} \autoref{fig:cross-domain} shows that cross-domain monofact rate can affect discovery in the presence of in-domain data. Under \emph{Mixed Domains}, materials discovery generally increases with the regulatory genomics monofact rate, although, unlike in \S\ref{subsec:results}, the relationship is not monotonic. Nevertheless, discovery tends to improve along with monofact rate, with the rate $1.0$ models slightly exceeding the \emph{In-Domain} baseline on both metrics. In the \emph{Cross-Domain} setting, however, absolute discovery rate drops substantially and the relationship with monofact rate disappears. These results therefore suggest that cross-domain monofacts can shape discovery when paired with in-domain data, but do not induce a domain-independent discovery capacity.

\section{Conclusion and Limitations}

\paragraph{Conclusion.} We study scientific discovery through the missing-mass perspective previously used to explain hallucination in language models~\citep{kalai2024calibrated}. Our analysis shows that monofact rate governs discovery capacity (\S\ref{sec:discovery}), and our experiments in materials science and regulatory genomics validate that increasing monofact rate increases discovery (\S\ref{sec:exp}). This relationship persists across different inference-time reasoning budgets (\S\ref{subsec:effort-ablation}), weakens as calibration error increases (\S\ref{subsec:calibration-error}), cannot be explained by training-set diversity alone (\S\ref{subsec:or}), and transfers across domains with in-domain support (\S\ref{subsec:cross-domain}). 

Our findings offer a new perspective on the long tail of scientific data. Prior work emphasizes challenges associated with it, including unreliable factual recall~\citep{kandpal2023large}, privacy and safety risks~\citep{shen2024languagebarrier}, and reduced distributional diversity~\citep{shumailov2024aimodels}. Our results, in contrast, reveal a potential benefit, as monofact rate signals that the training data has not exhausted the underlying distribution, leaving room for unseen outcomes to be discovered.

\paragraph{Limitations.} Our theoretical bounds depend on frontier parameters $\alpha$ and $\beta$, which are difficult to estimate in practice. This limits their applicability, although our central empirical findings concern the monofact rate and discovery relationship rather than the numerical values of bounds.

Notably, we measure discovery as exact recovery of targets that postdate the models' knowledge cutoffs (App.~\ref{app:novelty}). This is the standard offline test of discovery \citep{sourati2023accelerating, yang2025moosechem, riebesell2025framework, tshitoyan2019unsupervised}, but it remains conservative, since it might fail to credit alternative candidates that satisfy the constraints yet differ from the recorded target.

Future work should study \emph{monofact plasticity}: how monofact rate evolves across the training pipeline (from pretraining and midtraining to posttraining) and whether the effects of monofacts introduced at one stage persist, compound, or are overwritten by later training.

\section*{Acknowledgements}
We thank Mohd Zaki for his help with experiment design in materials discovery.
This work was supported in part by Apple, Defense Advance Research Projects Agency (DARPA)
under Contract No. HR001125C0304, ONR grant (N0001424-1-2089), JHU PURA (the Provost’s Undergraduate Research Award) and the JHU Pistritto Research Fellowship.
Any opinions, findings and conclusions or recommendations expressed in this material are those of the author(s) and do not necessarily reflect the views of DARPA.
We acknowledge the use of computational resources on the Johns Hopkins Data Science and AI Institute (DSAI) cluster.

\section*{AI Statement}
All ideas underlying the theoretical framework (\S\ref{sec:discovery}) originate with the authors, but Claude Fable 5.1 was used to check and revise all proofs. All experiments were proposed and designed by the authors and implemented by Claude and GPT models. Separately, Claude Opus 5 was used as a teacher model to generate reasoning traces for training, as described in App.~\ref{app:opus5-distill}.

\bibliography{conference}

\begin{thebibliography}{57}
\providecommand{\natexlab}[1]{#1}
\providecommand{\url}[1]{\texttt{#1}}
\expandafter\ifx\csname urlstyle\endcsname\relax
  \providecommand{\doi}[1]{doi: #1}\else
  \providecommand{\doi}{doi: \begingroup \urlstyle{rm}\Url}\fi

\bibitem[Aitchison(2024)]{aitchison2024overfit}
Laurence Aitchison.
\newblock Why you don't overfit, and don't need bayes if you only train for one epoch, 2024.
\newblock URL \url{https://arxiv.org/abs/2411.14478}.

\bibitem[Alp{\"o}ge(2026)]{alpoge2026jacobian}
Levent Alp{\"o}ge.
\newblock An explicit counterexample to the {J}acobian conjecture in dimension three.
\newblock Post on X, July 2026.
\newblock URL \url{https://x.com/__alpoge__/status/2079028340955197566}.
\newblock Found with Claude Fable 5; problem suggested by Akhil Mathew.

\bibitem[{Anthropic}(2026)]{opus5}
{Anthropic}.
\newblock Claude opus 5 system card.
\newblock Technical report, Anthropic, July 2026.
\newblock URL \url{https://www-cdn.anthropic.com/b514064af1408018e64b1ad24e7d5e75850b4ffd/Claude%20Opus%205%20System%20Card.pdf}.
\newblock Accessed: August 7, 2026.

\bibitem[Antunes et~al.(2024)Antunes, Butler, and Grau-Crespo]{antunes2024crystallm}
Luis~M. Antunes, Keith~T. Butler, and Ricardo Grau-Crespo.
\newblock Crystal structure generation with autoregressive large language modeling.
\newblock \emph{Nature Communications}, 15:\penalty0 10570, 2024.
\newblock \doi{10.1038/s41467-024-54639-7}.

\bibitem[Boiko et~al.(2023)Boiko, MacKnight, Kline, and Gomes]{boiko2023autonomous}
Daniil~A. Boiko, Robert MacKnight, Ben Kline, and Gabe Gomes.
\newblock Autonomous chemical research with large language models.
\newblock \emph{Nature}, 624:\penalty0 570--578, 2023.
\newblock \doi{10.1038/s41586-023-06792-0}.

\bibitem[Bran et~al.(2024)Bran, Cox, Schilter, Baldassari, White, and Schwaller]{bran2024augmenting}
Andres~M. Bran, Sam Cox, Oliver Schilter, Carlo Baldassari, Andrew~D. White, and Philippe Schwaller.
\newblock Augmenting large language models with chemistry tools.
\newblock \emph{Nature Machine Intelligence}, 6:\penalty0 525--535, 2024.
\newblock \doi{10.1038/s42256-024-00832-8}.

\bibitem[Brunton et~al.(2016)Brunton, Proctor, and Kutz]{brunton2016discovering}
Steven~L. Brunton, Joshua~L. Proctor, and J.~Nathan Kutz.
\newblock Discovering governing equations from data by sparse identification of nonlinear dynamical systems.
\newblock \emph{Proceedings of the National Academy of Sciences}, 113\penalty0 (15):\penalty0 3932--3937, 2016.
\newblock \doi{10.1073/pnas.1517384113}.

\bibitem[Carlini et~al.(2021)Carlini, Tramer, Wallace, Jagielski, Herbert-Voss, Lee, Roberts, Brown, Song, Erlingsson, Oprea, and Raffel]{carlini2021extracting}
Nicholas Carlini, Florian Tramer, Eric Wallace, Matthew Jagielski, Ariel Herbert-Voss, Katherine Lee, Adam Roberts, Tom Brown, Dawn Song, Ulfar Erlingsson, Alina Oprea, and Colin Raffel.
\newblock Extracting training data from large language models, 2021.
\newblock URL \url{https://arxiv.org/abs/2012.07805}.

\bibitem[Chen et~al.(2025)Chen, Asai, Zettlemoyer, Hajishirzi, and Brahman]{chen2025traintruthskillsbinary}
Tong Chen, Akari Asai, Luke Zettlemoyer, Hannaneh Hajishirzi, and Faeze Brahman.
\newblock Train for truth, keep the skills: Binary retrieval-augmented reward mitigates hallucinations, 2025.
\newblock URL \url{https://arxiv.org/abs/2510.17733}.

\bibitem[Cranmer et~al.(2020)Cranmer, Sanchez-Gonzalez, Battaglia, Xu, Cranmer, Spergel, and Ho]{cranmer2020discovering}
Miles Cranmer, Alvaro Sanchez-Gonzalez, Peter Battaglia, Rui Xu, Kyle Cranmer, David Spergel, and Shirley Ho.
\newblock Discovering symbolic models from deep learning with inductive biases.
\newblock In \emph{Advances in Neural Information Processing Systems}, volume~33, 2020.
\newblock URL \url{https://proceedings.neurips.cc/paper/2020/hash/c9f2f917078bd2db12f23c3b413d9cba-Abstract.html}.

\bibitem[Dalla-Torre et~al.(2025)Dalla-Torre, Gonzalez, Mendoza-Revilla, Lopez~Carranza, Grzywaczewski, Oteri, Dallago, Trop, de~Almeida, Sirelkhatim, Richard, Skwark, Beguir, Lopez, and Pierrot]{dallatorre2025nucleotide}
Hugo Dalla-Torre, Liam Gonzalez, Javier Mendoza-Revilla, Nicolas Lopez~Carranza, Adam~Henryk Grzywaczewski, Francesco Oteri, Christian Dallago, Evan Trop, Bernardo~P. de~Almeida, Hassan Sirelkhatim, Guillaume Richard, Marcin Skwark, Karim Beguir, Marie Lopez, and Thomas Pierrot.
\newblock Nucleotide transformer: building and evaluating robust foundation models for human genomics.
\newblock \emph{Nature Methods}, 22\penalty0 (2):\penalty0 287--297, 2025.
\newblock \doi{10.1038/s41592-024-02523-z}.

\bibitem[Das \& Srivastava(2026)Das and Srivastava]{das2026innovationcharacterizationhallucination}
Nishant~P. Das and Piyush Srivastava.
\newblock Innovation: An almost characterization of hallucination, 2026.
\newblock URL \url{https://arxiv.org/abs/2605.26808}.

\bibitem[Farquhar et~al.(2024)Farquhar, Kossen, Kuhn, and Gal]{Farquhar2024DetectingHI}
Sebastian Farquhar, Jannik Kossen, Lorenz Kuhn, and Yarin Gal.
\newblock Detecting hallucinations in large language models using semantic entropy.
\newblock \emph{Nature}, 630:\penalty0 625--630, 2024.
\newblock URL \url{https://api.semanticscholar.org/CorpusID:270615909}.

\bibitem[Gekhman et~al.(2024)Gekhman, Yona, Aharoni, Eyal, Feder, Reichart, and Herzig]{gekhman2024finetuning}
Zorik Gekhman, Gal Yona, Roee Aharoni, Matan Eyal, Amir Feder, Roi Reichart, and Jonathan Herzig.
\newblock Does fine-tuning {LLMs} on new knowledge encourage hallucinations?
\newblock In \emph{Proceedings of the 2024 Conference on Empirical Methods in Natural Language Processing}, 2024.

\bibitem[Gleick(2007)]{gleick2007isaac}
James Gleick.
\newblock \emph{Isaac Newton}.
\newblock Vintage, 2007.

\bibitem[Good(1953)]{good1953population}
I.~J. Good.
\newblock The population frequencies of species and the estimation of population parameters.
\newblock \emph{Biometrika}, 40\penalty0 (3/4):\penalty0 237--264, 1953.
\newblock ISSN 00063444, 14643510.
\newblock URL \url{http://www.jstor.org/stable/2333344}.

\bibitem[Gosai et~al.(2024)Gosai, Castro, Fuentes, Butts, Mouri, Alasoadura, Kales, Nguyen, Noche, Rao, Joy, Sabeti, Reilly, and Tewhey]{gosai2024machine}
Sager~J. Gosai, Rodrigo~I. Castro, Natalia Fuentes, John~C. Butts, Kousuke Mouri, Michael Alasoadura, Susan Kales, Thanh Thanh~L. Nguyen, Ramil~R. Noche, Arya~S. Rao, Mary~T. Joy, Pardis~C. Sabeti, Steven~K. Reilly, and Ryan Tewhey.
\newblock Machine-guided design of cell-type-targeting cis-regulatory elements.
\newblock \emph{Nature}, 634\penalty0 (8036):\penalty0 1211--1220, 2024.
\newblock \doi{10.1038/s41586-024-08070-z}.

\bibitem[Gottweis et~al.(2026)Gottweis, Weng, Daryin, Tu, Sirkovic, Myaskovsky, Glowaty, Weissenberger, Orlandi, Popovici, Palepu, Rong, Tanno, Saab, Zhang, Blum, Carroll, Kulkarni, Tomašev, Zverinski, Rendulic, Vedadi, Hasler, Rimanic, Boia, Budiselic, Feinstein, Bellaiche, Sheffer, Freyberg, Ratcliff, Bertolli, Chou, Hassidim, Gokturk, Vahdat, Guan, Dhillon, Vaishnav, Lee, Costa, Penadés, Peltz, Matias, Manyika, Hassabis, Xu, Kohli, Pawlosky, Karthikesalingam, and Natarajan]{gottweis2025coscientist}
Juraj Gottweis, Wei-Hung Weng, Alexander Daryin, Tao Tu, Petar Sirkovic, Artiom Myaskovsky, Grzegorz Glowaty, Felix Weissenberger, Alessio Orlandi, Dan Popovici, Anil Palepu, Keran Rong, Ryutaro Tanno, Khaled Saab, Fan Zhang, Jacob Blum, Andrew Carroll, Kavita Kulkarni, Nenad Tomašev, Dina Zverinski, Ivor Rendulic, Elahe Vedadi, Florian Hasler, Luka Rimanic, Marina Boia, Ivan Budiselic, Ben Feinstein, Mathias Bellaiche, Tom Sheffer, Jan Freyberg, Jeremy Ratcliff, Ottavia Bertolli, Katherine Chou, Avinatan Hassidim, Burak Gokturk, Amin Vahdat, Yuan Guan, Vikram Dhillon, Eeshit~Dhaval Vaishnav, Byron Lee, Tiago R.~D. Costa, José~R. Penadés, Gary Peltz, Yossi Matias, James Manyika, Demis Hassabis, Yunhan Xu, Pushmeet Kohli, Annalisa Pawlosky, Alan Karthikesalingam, and Vivek Natarajan.
\newblock Accelerating scientific discovery with co-scientist.
\newblock \emph{Nature}, 655\penalty0 (8122):\penalty0 487--496, May 2026.
\newblock ISSN 1476-4687.
\newblock \doi{10.1038/s41586-026-10644-y}.
\newblock URL \url{http://dx.doi.org/10.1038/s41586-026-10644-y}.

\bibitem[Guo et~al.(2017)Guo, Pleiss, Sun, and Weinberger]{guo2017calibration}
Chuan Guo, Geoff Pleiss, Yu~Sun, and Kilian~Q. Weinberger.
\newblock On calibration of modern neural networks, 2017.
\newblock URL \url{https://arxiv.org/abs/1706.04599}.

\bibitem[Jablonka et~al.(2024)Jablonka, Schwaller, Ortega-Guerrero, and Smit]{jablonka2024predictive}
Kevin~Maik Jablonka, Philippe Schwaller, Andres Ortega-Guerrero, and Berend Smit.
\newblock Leveraging large language models for predictive chemistry.
\newblock \emph{Nature Machine Intelligence}, 6:\penalty0 161--169, 2024.
\newblock \doi{10.1038/s42256-023-00788-1}.

\bibitem[Jain et~al.(2013)Jain, Ong, Hautier, Chen, Richards, Dacek, Cholia, Gunter, Skinner, Ceder, and Persson]{jain2013materials}
Anubhav Jain, Shyue~Ping Ong, Geoffroy Hautier, Wei Chen, William~Davidson Richards, Stephen Dacek, Shreyas Cholia, Dan Gunter, David Skinner, Gerbrand Ceder, and Kristin~A. Persson.
\newblock Commentary: The materials project: A materials genome approach to accelerating materials innovation.
\newblock \emph{APL Materials}, 1\penalty0 (1):\penalty0 011002, 2013.
\newblock \doi{10.1063/1.4812323}.

\bibitem[Ji et~al.(2023)Ji, Lee, Frieske, Yu, Su, Xu, Ishii, Bang, Madotto, and Fung]{ji2023survey}
Ziwei Ji, Nayeon Lee, Rita Frieske, Tiezheng Yu, Dan Su, Yan Xu, Etsuko Ishii, Ye~Jin Bang, Andrea Madotto, and Pascale Fung.
\newblock Survey of hallucination in natural language generation.
\newblock \emph{ACM Computing Surveys}, 55\penalty0 (12):\penalty0 1--38, March 2023.
\newblock ISSN 1557-7341.
\newblock \doi{10.1145/3571730}.
\newblock URL \url{http://dx.doi.org/10.1145/3571730}.

\bibitem[Kalai \& Vempala(2024)Kalai and Vempala]{kalai2024calibrated}
Adam~Tauman Kalai and Santosh~S. Vempala.
\newblock Calibrated language models must hallucinate.
\newblock In \emph{Proceedings of the 56th Annual ACM Symposium on Theory of Computing}, pp.\  160--171. ACM, 2024.
\newblock \doi{10.1145/3618260.3649777}.

\bibitem[Kandpal et~al.(2023)Kandpal, Deng, Roberts, Wallace, and Raffel]{kandpal2023large}
Nikhil Kandpal, Haikang Deng, Adam Roberts, Eric Wallace, and Colin Raffel.
\newblock Large language models struggle to learn long-tail knowledge.
\newblock In Andreas Krause, Emma Brunskill, Kyunghyun Cho, Barbara Engelhardt, Sivan Sabato, and Jonathan Scarlett (eds.), \emph{Proceedings of the 40th International Conference on Machine Learning}, volume 202 of \emph{Proceedings of Machine Learning Research}, pp.\  15696--15707. PMLR, 23--29 Jul 2023.
\newblock URL \url{https://proceedings.mlr.press/v202/kandpal23a.html}.

\bibitem[Kuhn et~al.(2023)Kuhn, Gal, and Farquhar]{kuhn2023semantic}
Lorenz Kuhn, Yarin Gal, and Sebastian Farquhar.
\newblock Semantic uncertainty: Linguistic invariances for uncertainty estimation in natural language generation.
\newblock In \emph{The Eleventh International Conference on Learning Representations}, 2023.

\bibitem[Kusne et~al.(2020)Kusne, Yu, Wu, Zhang, Hattrick-Simpers, DeCost, Sarker, Oses, Toher, Curtarolo, Davydov, Agarwal, Bendersky, Li, Mehta, and Takeuchi]{kusne2020closedloop}
A.~Gilad Kusne, Heshan Yu, Changming Wu, Huairuo Zhang, Jason Hattrick-Simpers, Brian DeCost, Suchismita Sarker, Corey Oses, Cormac Toher, Stefano Curtarolo, Albert~V. Davydov, Ritesh Agarwal, Leonid~A. Bendersky, Mo~Li, Apurva Mehta, and Ichiro Takeuchi.
\newblock On-the-fly closed-loop materials discovery via bayesian active learning.
\newblock \emph{Nature Communications}, 11:\penalty0 5966, 2020.
\newblock \doi{10.1038/s41467-020-19597-w}.

\bibitem[Lin(1991)]{lin1991divergence}
Jianhua Lin.
\newblock Divergence measures based on the shannon entropy.
\newblock \emph{IEEE Transactions on Information Theory}, 37\penalty0 (1):\penalty0 145--151, 1991.

\bibitem[Lin et~al.(2023)Lin, Akin, Rao, Hie, Zhu, Lu, Smetanin, Verkuil, Kabeli, Shmueli, Dos Santos~Costa, Fazel-Zarandi, Sercu, Candido, and Rives]{lin2023evolutionary}
Zeming Lin, Halil Akin, Roshan Rao, Brian Hie, Zhongkai Zhu, Wenting Lu, Nikita Smetanin, Robert Verkuil, Ori Kabeli, Yaniv Shmueli, Allan Dos Santos~Costa, Maryam Fazel-Zarandi, Tom Sercu, Salvatore Candido, and Alexander Rives.
\newblock Evolutionary-scale prediction of atomic-level protein structure with a language model.
\newblock \emph{Science}, 379\penalty0 (6637):\penalty0 1123--1130, 2023.
\newblock \doi{10.1126/science.ade2574}.

\bibitem[Mallen et~al.(2023)Mallen, Asai, Zhong, Das, Khashabi, and Hajishirzi]{mallen2023trust}
Alex Mallen, Akari Asai, Victor Zhong, Rajarshi Das, Daniel Khashabi, and Hannaneh Hajishirzi.
\newblock When not to trust language models: Investigating effectiveness of parametric and non-parametric memories, 2023.
\newblock URL \url{https://arxiv.org/abs/2212.10511}.

\bibitem[Merchant et~al.(2023)Merchant, Batzner, Schoenholz, Aykol, Cheon, and Cubuk]{merchant2023scaling}
Amil Merchant, Simon Batzner, Samuel~S. Schoenholz, Muratahan Aykol, Gowoon Cheon, and Ekin~Dogus Cubuk.
\newblock Scaling deep learning for materials discovery.
\newblock \emph{Nature}, 624:\penalty0 80--85, 2023.
\newblock \doi{10.1038/s41586-023-06735-9}.

\bibitem[Meredig et~al.(2018)Meredig, Antono, Church, Hutchinson, Ling, Paradiso, Blaiszik, Foster, Gibbons, Hattrick-Simpers, Mehta, and Ward]{meredig2018extrapolation}
Bryce Meredig, Erin Antono, Carena Church, Maxwell Hutchinson, Julia Ling, Sean Paradiso, Ben Blaiszik, Ian Foster, Brenna Gibbons, Jason Hattrick-Simpers, Apurva Mehta, and Logan Ward.
\newblock Can machine learning identify the next high-temperature superconductor? examining extrapolation performance for materials discovery.
\newblock \emph{Molecular Systems Design \& Engineering}, 3:\penalty0 819--825, 2018.
\newblock \doi{10.1039/C8ME00012C}.

\bibitem[Miao \& Kearns(2026)Miao and Kearns]{miao2025hallucination}
Miranda~Muqing Miao and Michael Kearns.
\newblock Hallucination, monofacts, and miscalibration: An empirical investigation, 2026.
\newblock URL \url{https://arxiv.org/abs/2502.08666}.

\bibitem[Min et~al.(2023)Min, Krishna, Lyu, Lewis, Yih, Koh, Iyyer, Zettlemoyer, and Hajishirzi]{min-etal-2023-factscore}
Sewon Min, Kalpesh Krishna, Xinxi Lyu, Mike Lewis, Wen-tau Yih, Pang Koh, Mohit Iyyer, Luke Zettlemoyer, and Hannaneh Hajishirzi.
\newblock {FA}ct{S}core: Fine-grained atomic evaluation of factual precision in long form text generation.
\newblock In Houda Bouamor, Juan Pino, and Kalika Bali (eds.), \emph{Proceedings of the 2023 Conference on Empirical Methods in Natural Language Processing}, pp.\  12076--12100, Singapore, December 2023. Association for Computational Linguistics.
\newblock \doi{10.18653/v1/2023.emnlp-main.741}.
\newblock URL \url{https://aclanthology.org/2023.emnlp-main.741/}.

\bibitem[Nakkiran et~al.(2026)Nakkiran, Bradley, Golinski, Ndiaye, Kirchhof, and Williamson]{nakkiran2026trained}
Preetum Nakkiran, Arwen Bradley, Adam Golinski, Eugene Ndiaye, Michael Kirchhof, and Sinead Williamson.
\newblock Trained on tokens, calibrated on concepts: The emergence of semantic calibration in llms.
\newblock In \emph{International Conference on Learning Representations}, 2026.

\bibitem[Newman(2005)]{newman2005power}
M.~E.~J. Newman.
\newblock Power laws, pareto distributions and zipf's law.
\newblock \emph{Contemporary Physics}, 46\penalty0 (5):\penalty0 323--351, 2005.
\newblock \doi{10.1080/00107510500052444}.

\bibitem[Noorani et~al.(2025)Noorani, Kiyani, Pappas, and Hassani]{noorani2025conformal}
Sima Noorani, Shayan Kiyani, George Pappas, and Hamed Hassani.
\newblock Conformal prediction beyond the seen: A missing mass perspective for uncertainty quantification in generative models, 2025.
\newblock URL \url{https://arxiv.org/abs/2506.05497}.

\bibitem[OpenAI(2025)]{openai2025gptoss120bgptoss20bmodel}
OpenAI.
\newblock gpt-oss-120b \& gpt-oss-20b model card, 2025.
\newblock URL \url{https://arxiv.org/abs/2508.10925}.

\bibitem[{OpenAI}(2026{\natexlab{a}})]{openai2026navierstokes}
{OpenAI}.
\newblock On the navier--stokes millennium prize problem, September 2026{\natexlab{a}}.
\newblock OpenAI Research.

\bibitem[{OpenAI}(2026{\natexlab{b}})]{openai2026scientificcollaborator}
{OpenAI}.
\newblock {AI as a Scientific Collaborator: From biology to black holes, ChatGPT is accelerating research}.
\newblock Technical report, OpenAI, January 2026{\natexlab{b}}.
\newblock URL \url{https://cdn.openai.com/pdf/f4b4a5da-b2de-418d-9fcd-6b293e9dc157/oai_ai-as-a-scientific-collaborator_jan-2026.pdf}.

\bibitem[{OpenAI}(2026{\natexlab{c}})]{openai2026unitdistance}
{OpenAI}.
\newblock An {OpenAI} model has disproved a central conjecture in discrete geometry.
\newblock OpenAI Research, May 2026{\natexlab{c}}.
\newblock URL \url{https://openai.com/index/model-disproves-discrete-geometry-conjecture/}.

\bibitem[Pareto(1896)]{pareto1896cours}
Vilfredo Pareto.
\newblock \emph{Cours d'{\'e}conomie politique profess{\'e} {\`a} l'Universit{\'e} de Lausanne}, volume~1.
\newblock F. Rouge, Lausanne, 1896.

\bibitem[Razeghi et~al.(2022)Razeghi, Logan~IV, Gardner, and Singh]{razeghi2022impact}
Yasaman Razeghi, Robert~L. Logan~IV, Matt Gardner, and Sameer Singh.
\newblock Impact of pretraining term frequencies on few-shot reasoning, 2022.
\newblock URL \url{https://arxiv.org/abs/2202.07206}.

\bibitem[Riebesell et~al.(2025)Riebesell, Goodall, Benner, Chiang, Deng, Ceder, Asta, Lee, Jain, and Persson]{riebesell2025framework}
Janosh Riebesell, Rhys E.~A. Goodall, Philipp Benner, Yuan Chiang, Bowen Deng, Gerbrand Ceder, Mark Asta, Alpha~A. Lee, Anubhav Jain, and Kristin~A. Persson.
\newblock A framework to evaluate machine learning crystal stability predictions.
\newblock \emph{Nature Machine Intelligence}, 7\penalty0 (6):\penalty0 836--847, 2025.
\newblock \doi{10.1038/s42256-025-01055-1}.

\bibitem[Romera-Paredes et~al.(2024)Romera-Paredes, Barekatain, Novikov, Balog, Kumar, Dupont, Ruiz, Ellenberg, Wang, Fawzi, Kohli, and Fawzi]{romeraparedes2024mathematical}
Bernardino Romera-Paredes, Mohammadamin Barekatain, Alexander Novikov, Matej Balog, M.~Pawan Kumar, Emilien Dupont, Francisco J.~R. Ruiz, Jordan~S. Ellenberg, Pengming Wang, Omar Fawzi, Pushmeet Kohli, and Alhussein Fawzi.
\newblock Mathematical discoveries from program search with large language models.
\newblock \emph{Nature}, 625\penalty0 (7995):\penalty0 468--475, 2024.
\newblock \doi{10.1038/s41586-023-06924-6}.

\bibitem[Shabani et~al.(2010)Shabani, Khatib, and Ebadi]{shabani2010vygotsky}
Karim Shabani, Mohamad Khatib, and Saman Ebadi.
\newblock Vygotsky's zone of proximal development: Instructional implications and teachers' professional development.
\newblock \emph{English Language Teaching}, 3:\penalty0 237--248, 2010.
\newblock URL \url{https://api.semanticscholar.org/CorpusID:38382898}.

\bibitem[Shen et~al.(2024)Shen, Tan, Chen, Chen, Zhang, Xu, Zheng, Koehn, and Khashabi]{shen2024languagebarrier}
Lingfeng Shen, Weiting Tan, Sihao Chen, Yunmo Chen, Jingyu Zhang, Haoran Xu, Boyuan Zheng, Philipp Koehn, and Daniel Khashabi.
\newblock The language barrier: Dissecting safety challenges of {LLMs} in multilingual contexts, 2024.
\newblock URL \url{https://arxiv.org/abs/2401.13136}.

\bibitem[Shumailov et~al.(2024)Shumailov, Shumaylov, Zhao, Papernot, Anderson, and Gal]{shumailov2024aimodels}
Ilia Shumailov, Zakhar Shumaylov, Yiren Zhao, Nicolas Papernot, Ross Anderson, and Yarin Gal.
\newblock Ai models collapse when trained on recursively generated data.
\newblock \emph{Nature}, 631\penalty0 (8022):\penalty0 755--759, 2024.
\newblock \doi{10.1038/s41586-024-07566-y}.
\newblock URL \url{https://doi.org/10.1038/s41586-024-07566-y}.

\bibitem[Sourati \& Evans(2023)Sourati and Evans]{sourati2023accelerating}
Jamshid Sourati and James Evans.
\newblock Accelerating science with human-aware artificial intelligence, 2023.
\newblock URL \url{https://arxiv.org/abs/2306.01495}.

\bibitem[Sun et~al.(2024)Sun, Xu, Zha, Liu, and Dong]{sun2024head}
Kai Sun, Yifan~Ethan Xu, Hanwen Zha, Yue Liu, and Xin~Luna Dong.
\newblock Head-to-tail: How knowledgeable are large language models (llms)? a.k.a. will llms replace knowledge graphs?, 2024.
\newblock URL \url{https://arxiv.org/abs/2308.10168}.

\bibitem[Team et~al.(2025)Team, Kamath, Ferret, Pathak, Vieillard, Merhej, Perrin, Matejovicova, Ramé, Rivière, Rouillard, Mesnard, Cideron, bastien Grill, Ramos, Yvinec, Casbon, Pot, Penchev, Liu, Visin, Kenealy, Beyer, Zhai, Tsitsulin, Busa-Fekete, Feng, Sachdeva, Coleman, Gao, Mustafa, Barr, Parisotto, Tian, Eyal, Cherry, Peter, Sinopalnikov, Bhupatiraju, Agarwal, Kazemi, Malkin, Kumar, Vilar, Brusilovsky, Luo, Steiner, Friesen, Sharma, Sharma, Gilady, Goedeckemeyer, Saade, Feng, Kolesnikov, Bendebury, Abdagic, Vadi, György, Pinto, Das, Bapna, Miech, Yang, Paterson, Shenoy, Chakrabarti, Piot, Wu, Shahriari, Petrini, Chen, Lan, Choquette-Choo, Carey, Brick, Deutsch, Eisenbud, Cattle, Cheng, Paparas, Sreepathihalli, Reid, Tran, Zelle, Noland, Huizenga, Kharitonov, Liu, Amirkhanyan, Cameron, Hashemi, Klimczak-Plucińska, Singh, Mehta, Lehri, Hazimeh, Ballantyne, Szpektor, Nardini, Pouget-Abadie, Chan, Stanton, Wieting, Lai, Orbay, Fernandez, Newlan, yeong Ji, Singh, Black, Yu, Hui, Vodrahalli, Greff, Qiu,
  Valentine, Coelho, Ritter, Hoffman, Watson, Chaturvedi, Moynihan, Ma, Babar, Noy, Byrd, Roy, Momchev, Chauhan, Sachdeva, Bunyan, Botarda, Caron, Rubenstein, Culliton, Schmid, Sessa, Xu, Stanczyk, Tafti, Shivanna, Wu, Pan, Rokni, Willoughby, Vallu, Mullins, Jerome, Smoot, Girgin, Iqbal, Reddy, Sheth, Põder, Bhatnagar, Panyam, Eiger, Zhang, Liu, Yacovone, Liechty, Kalra, Evci, Misra, Roseberry, Feinberg, Kolesnikov, Han, Kwon, Chen, Chow, Zhu, Wei, Egyed, Cotruta, Giang, Kirk, Rao, Black, Babar, Lo, Moreira, Martins, Sanseviero, Gonzalez, Gleicher, Warkentin, Mirrokni, Senter, Collins, Barral, Ghahramani, Hadsell, Matias, Sculley, Petrov, Fiedel, Shazeer, Vinyals, Dean, Hassabis, Kavukcuoglu, Farabet, Buchatskaya, Alayrac, Anil, Lepikhin, Borgeaud, Bachem, Joulin, Andreev, Hardin, Dadashi, and Hussenot]{gemmateam2025gemma3technicalreport}
Gemma Team, Aishwarya Kamath, Johan Ferret, Shreya Pathak, Nino Vieillard, Ramona Merhej, Sarah Perrin, Tatiana Matejovicova, Alexandre Ramé, Morgane Rivière, Louis Rouillard, Thomas Mesnard, Geoffrey Cideron, Jean bastien Grill, Sabela Ramos, Edouard Yvinec, Michelle Casbon, Etienne Pot, Ivo Penchev, Gaël Liu, Francesco Visin, Kathleen Kenealy, Lucas Beyer, Xiaohai Zhai, Anton Tsitsulin, Robert Busa-Fekete, Alex Feng, Noveen Sachdeva, Benjamin Coleman, Yi~Gao, Basil Mustafa, Iain Barr, Emilio Parisotto, David Tian, Matan Eyal, Colin Cherry, Jan-Thorsten Peter, Danila Sinopalnikov, Surya Bhupatiraju, Rishabh Agarwal, Mehran Kazemi, Dan Malkin, Ravin Kumar, David Vilar, Idan Brusilovsky, Jiaming Luo, Andreas Steiner, Abe Friesen, Abhanshu Sharma, Abheesht Sharma, Adi~Mayrav Gilady, Adrian Goedeckemeyer, Alaa Saade, Alex Feng, Alexander Kolesnikov, Alexei Bendebury, Alvin Abdagic, Amit Vadi, András György, André~Susano Pinto, Anil Das, Ankur Bapna, Antoine Miech, Antoine Yang, Antonia Paterson, Ashish
  Shenoy, Ayan Chakrabarti, Bilal Piot, Bo~Wu, Bobak Shahriari, Bryce Petrini, Charlie Chen, Charline~Le Lan, Christopher~A. Choquette-Choo, CJ~Carey, Cormac Brick, Daniel Deutsch, Danielle Eisenbud, Dee Cattle, Derek Cheng, Dimitris Paparas, Divyashree~Shivakumar Sreepathihalli, Doug Reid, Dustin Tran, Dustin Zelle, Eric Noland, Erwin Huizenga, Eugene Kharitonov, Frederick Liu, Gagik Amirkhanyan, Glenn Cameron, Hadi Hashemi, Hanna Klimczak-Plucińska, Harman Singh, Harsh Mehta, Harshal~Tushar Lehri, Hussein Hazimeh, Ian Ballantyne, Idan Szpektor, Ivan Nardini, Jean Pouget-Abadie, Jetha Chan, Joe Stanton, John Wieting, Jonathan Lai, Jordi Orbay, Joseph Fernandez, Josh Newlan, Ju~yeong Ji, Jyotinder Singh, Kat Black, Kathy Yu, Kevin Hui, Kiran Vodrahalli, Klaus Greff, Linhai Qiu, Marcella Valentine, Marina Coelho, Marvin Ritter, Matt Hoffman, Matthew Watson, Mayank Chaturvedi, Michael Moynihan, Min Ma, Nabila Babar, Natasha Noy, Nathan Byrd, Nick Roy, Nikola Momchev, Nilay Chauhan, Noveen Sachdeva, Oskar
  Bunyan, Pankil Botarda, Paul Caron, Paul~Kishan Rubenstein, Phil Culliton, Philipp Schmid, Pier~Giuseppe Sessa, Pingmei Xu, Piotr Stanczyk, Pouya Tafti, Rakesh Shivanna, Renjie Wu, Renke Pan, Reza Rokni, Rob Willoughby, Rohith Vallu, Ryan Mullins, Sammy Jerome, Sara Smoot, Sertan Girgin, Shariq Iqbal, Shashir Reddy, Shruti Sheth, Siim Põder, Sijal Bhatnagar, Sindhu~Raghuram Panyam, Sivan Eiger, Susan Zhang, Tianqi Liu, Trevor Yacovone, Tyler Liechty, Uday Kalra, Utku Evci, Vedant Misra, Vincent Roseberry, Vlad Feinberg, Vlad Kolesnikov, Woohyun Han, Woosuk Kwon, Xi~Chen, Yinlam Chow, Yuvein Zhu, Zichuan Wei, Zoltan Egyed, Victor Cotruta, Minh Giang, Phoebe Kirk, Anand Rao, Kat Black, Nabila Babar, Jessica Lo, Erica Moreira, Luiz~Gustavo Martins, Omar Sanseviero, Lucas Gonzalez, Zach Gleicher, Tris Warkentin, Vahab Mirrokni, Evan Senter, Eli Collins, Joelle Barral, Zoubin Ghahramani, Raia Hadsell, Yossi Matias, D.~Sculley, Slav Petrov, Noah Fiedel, Noam Shazeer, Oriol Vinyals, Jeff Dean, Demis Hassabis,
  Koray Kavukcuoglu, Clement Farabet, Elena Buchatskaya, Jean-Baptiste Alayrac, Rohan Anil, Dmitry Lepikhin, Sebastian Borgeaud, Olivier Bachem, Armand Joulin, Alek Andreev, Cassidy Hardin, Robert Dadashi, and Léonard Hussenot.
\newblock Gemma 3 technical report, 2025.
\newblock URL \url{https://arxiv.org/abs/2503.19786}.

\bibitem[Tshitoyan et~al.(2019)Tshitoyan, Dagdelen, Weston, Dunn, Rong, Kononova, Persson, Ceder, and Jain]{tshitoyan2019unsupervised}
Vahe Tshitoyan, John Dagdelen, Leigh Weston, Alexander Dunn, Ziqin Rong, Olga Kononova, Kristin~A. Persson, Gerbrand Ceder, and Anubhav Jain.
\newblock Unsupervised word embeddings capture latent knowledge from materials science literature.
\newblock \emph{Nature}, 571\penalty0 (7763):\penalty0 95--98, 2019.
\newblock \doi{10.1038/s41586-019-1335-8}.

\bibitem[Udrescu \& Tegmark(2020)Udrescu and Tegmark]{udrescu2020aifeynman}
Silviu-Marian Udrescu and Max Tegmark.
\newblock {AI Feynman}: A physics-inspired method for symbolic regression.
\newblock \emph{Science Advances}, 6\penalty0 (16):\penalty0 eaay2631, 2020.
\newblock \doi{10.1126/sciadv.aay2631}.

\bibitem[Xue et~al.(2023)Xue, Fu, Zhou, Zheng, and You]{xue2023repeat}
Fuzhao Xue, Yao Fu, Wangchunshu Zhou, Zangwei Zheng, and Yang You.
\newblock To repeat or not to repeat: Insights from scaling {LLM} under token-crisis, 2023.
\newblock URL \url{https://arxiv.org/abs/2305.13230}.

\bibitem[Yang et~al.(2025)Yang, Liu, Gao, Xie, Li, Ouyang, Poria, Cambria, and Zhou]{yang2025moosechem}
Zonglin Yang, Wanhao Liu, Ben Gao, Tong Xie, Yuqiang Li, Wanli Ouyang, Soujanya Poria, Erik Cambria, and Dongzhan Zhou.
\newblock {MOOSE-Chem}: Large language models for rediscovering unseen chemistry scientific hypotheses, 2025.
\newblock URL \url{https://arxiv.org/abs/2410.07076}.

\bibitem[Zeni et~al.(2025)Zeni, Pinsler, Z{\"u}gner, Fowler, Horton, Fu, Wang, Shysheya, Crabb{\'e}, Ueda, Sordillo, Sun, Smith, Nguyen, Schulz, Lewis, Huang, Lu, Zhou, Yang, Hao, Li, Yang, Li, Tomioka, and Xie]{zeni2025mattergen}
Claudio Zeni, Robert Pinsler, Daniel Z{\"u}gner, Andrew Fowler, Matthew Horton, Xiang Fu, Zilong Wang, Aliaksandra Shysheya, Jonathan Crabb{\'e}, Shoko Ueda, Roberto Sordillo, Lixin Sun, Jake Smith, Bichlien Nguyen, Hannes Schulz, Sarah Lewis, Chin-Wei Huang, Ziheng Lu, Yichi Zhou, Han Yang, Hongxia Hao, Jielan Li, Chunlei Yang, Wenjie Li, Ryota Tomioka, and Tian Xie.
\newblock A generative model for inorganic materials design.
\newblock \emph{Nature}, 639:\penalty0 624--632, 2025.
\newblock \doi{10.1038/s41586-025-08628-5}.

\bibitem[Zhang et~al.(2024)Zhang, Ding, Lyv, Wang, Yin, Zhang, Yu, Wang, Li, Xiang, Feng, Zhuang, Wang, Qin, Zhang, Zhang, Cui, Huang, Yan, Xu, Chen, Li, Fan, Xing, and Chen]{zhang2024scientificllms}
Qiang Zhang, Keyang Ding, Tianwen Lyv, Xinda Wang, Qingyu Yin, Yiwen Zhang, Jing Yu, Yuhao Wang, Xiaotong Li, Zhuoyi Xiang, Kehua Feng, Xiang Zhuang, Zeyuan Wang, Ming Qin, Mengyao Zhang, Jinlu Zhang, Jiyu Cui, Tao Huang, Pengju Yan, Renjun Xu, Hongyang Chen, Xiaolin Li, Xiaohui Fan, Huabin Xing, and Huajun Chen.
\newblock Scientific large language models: A survey on biological \& chemical domains.
\newblock \emph{arXiv preprint arXiv:2401.14656}, 2024.
\newblock URL \url{https://arxiv.org/abs/2401.14656}.

\bibitem[Zipf(1949)]{zipf1949human}
George~Kingsley Zipf.
\newblock \emph{Human Behavior and the Principle of Least Effort: An Introduction to Human Ecology}.
\newblock Addison-Wesley Press, Cambridge, MA, 1949.

\end{thebibliography}
\bibliographystyle{conference}
\newpage
\appendix

\section{Related Work}
\label{sec:prior-work}
\paragraph{Long-tail distributions in natural data.} 
Long-tailed frequency distributions, in which a few patterns are common while many are rare, have long been studied through Pareto and Zipf laws~\citep{pareto1896cours,zipf1949human,newman2005power}. In the context of LLMs, prior work shows that language models acquire knowledge unevenly across the frequency distribution of their pretraining data, with factual recall declining substantially for infrequently observed knowledge \citep{razeghi2022impact,kandpal2023large,mallen2023trust}. This frequency dependence persists with scale, as larger models continue to exhibit significant performance gaps between head and tail knowledge \citep{mallen2023trust,sun2024head}. Beyond factual recall, long-tail data raise concerns about privacy~\citep{carlini2021extracting}, safety~\citep{shen2024languagebarrier}, and the loss of rare patterns during recursive training~\citep{shumailov2024aimodels}. Our work considers a complementary aspect of the long-tail data by focusing on \emph{monofacts}: facts that occur exactly once and therefore occupy its extreme end. While prior work primarily characterizes extreme rarity as a challenge for model learning, our results reveal a positive association between monofacts and discovery, suggesting that long-tail data can also be beneficial.

\paragraph{Hallucination.}
Hallucination in language models is commonly studied as the generation of fluent but false statements, with prior work focused on its measurement~\citep{min-etal-2023-factscore}, detection~\citep{Farquhar2024DetectingHI}, and mitigation~\citep{ji2023survey,chen2025traintruthskillsbinary}. WWe build directly on a complementary line of work relating hallucination to calibration, monofact rate, and the missing probability mass~\citep{kalai2024calibrated,miao2025hallucination,das2026innovationcharacterizationhallucination}. Relatedly, \citet{gekhman2024finetuning} show that LLMs learn fine-tuning examples containing new knowledge more slowly than those consistent with pretraining, and that once learned, such examples linearly increase the model's tendency to hallucinate. We provide a detailed account of this framework~\citep{kalai2024calibrated} in \S\ref{sec:background} and extend it from characterizing unseen falsehoods (hallucinations) to characterizing unseen truths (discoveries).

\paragraph{LLMs for scientific discovery.}
Language models and machine learning systems increasingly support scientific discovery across a wide range of scientific domains~\citep{zhang2024scientificllms}. In biology, foundation models support protein structure prediction~\citep{lin2023evolutionary} and genomic representation learning and functional prediction~\citep{dallatorre2025nucleotide}. In chemistry and materials science, generative models support property prediction~\citep{jablonka2024predictive}, produce previously unobserved crystal structures~\citep{antunes2024crystallm}, and design materials that satisfy specified property constraints~\citep{zeni2025mattergen}. In mathematics and computer science, LLM-guided search identifies new constructions and algorithms~\citep{romeraparedes2024mathematical}, while recent frontier models contribute directly to open Erd\H{o}s problems~\citep{openai2026scientificcollaborator} and generate externally verified counterexamples to longstanding conjectures~\citep{openai2026unitdistance}. More generally, LLM-based agents operate over the literature, simulations, domain-specific tools, and experimental workflows to formulate and evaluate hypotheses~\citep{boiko2023autonomous,bran2024augmenting,gottweis2025coscientist}. 

Recent work also studies the conditions under which machine learning systems produce useful discoveries. Symbolic scientific discovery shows that extrapolation can depend on exploitable structure such as sparsity~\citep{brunton2016discovering}, symmetry, separability, and compositionality~\citep{udrescu2020aifeynman}, as well as domain-specific inductive biases~\citep{cranmer2020discovering}. More broadly, extrapolative performance depends on how observed data relate to unexplored regions of the scientific space~\citep{meredig2018extrapolation}, while active learning uses uncertainty to direct computation and experiments toward informative candidates~\citep{kusne2020closedloop}. Together, these works suggest that discovery depends not only on generating beyond observed data but also on whether the unseen region contains learnable structure. Our work gives this intuition a statistical account by connecting generation beyond the observed sample to missing mass and characterizing when probability assigned to unseen outputs yields discoveries rather than hallucinations.

\section{Complementary Analyses}

\subsection{Assessment of Calibration}
\label{app:calibration}

We assess whether improved calibration could explain the increase in discovery with monofact rate in materials discovery. For each input, we estimate formula-level probabilities from the empirical frequencies of formulas among sampled generations from \textbf{\texttt{pass@16}} evaluation. We measure expected calibration error (ECE) using 10 equal-width confidence bins on the test set.

\begin{figure}[h]
    \centering
    \begin{subfigure}{0.49\linewidth}
        \centering
        \includegraphics[width=\linewidth]{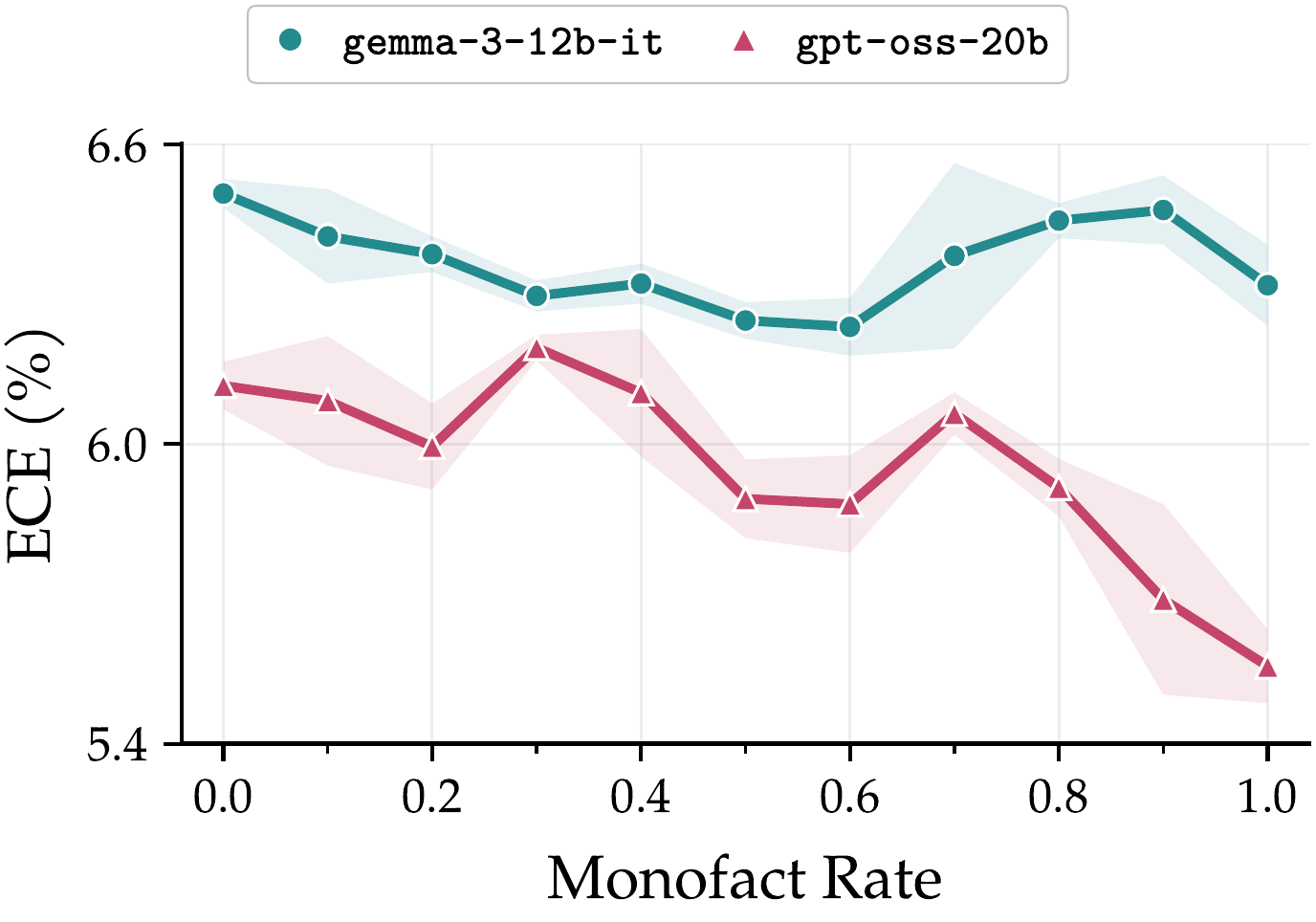}
        \caption{ECE across monofact rates for \texttt{gpt-oss-20b} and \texttt{gemma-3-12b-it}.}
        \label{fig:calibration-models}
    \end{subfigure}\hfill
    \begin{subfigure}{0.49\linewidth}
        \centering
        \includegraphics[width=\linewidth]{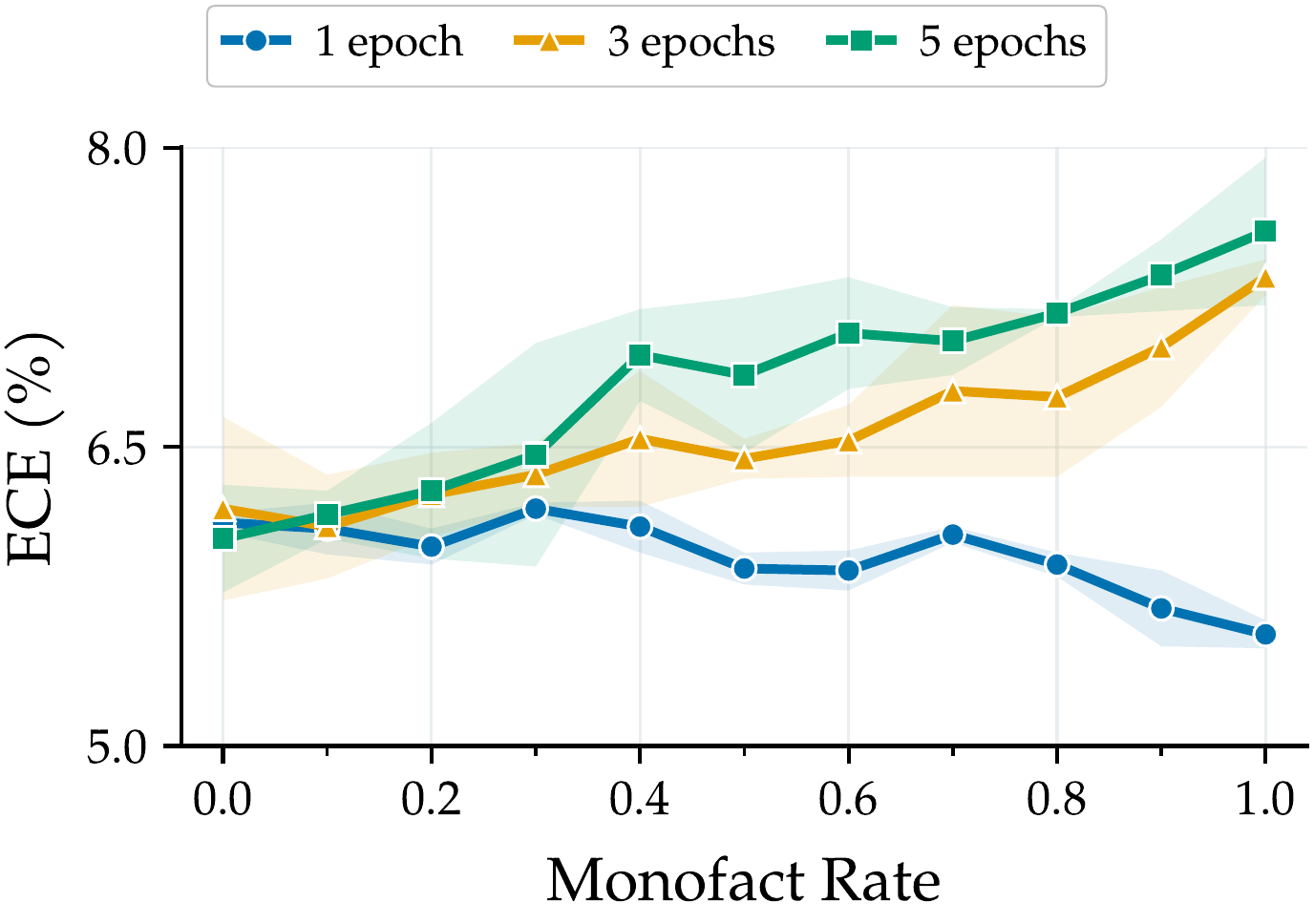}
        \caption{ECE across different training epochs for \texttt{gpt-oss-20b}.}
        \label{fig:calibration-epochs}
    \end{subfigure}
    \caption{\textbf{Calibration in materials science across models and training epochs.} Calibration error remains flat after a single training epoch but increases across monofact rates with additional training epochs. Shaded regions denote standard error.}
    \label{fig:calibration}
\end{figure}

\autoref{fig:calibration-models} shows that calibration remains relatively stable under one-epoch training across monofact rates for both model families, with a modest decrease in ECE for \textbf{\texttt{gpt-oss-20b}}. Thus, the increasing discovery trend cannot be attributed primarily to improved calibration.  In contrast, \autoref{fig:calibration-epochs} shows that this stability does not hold under repeated training. With three and five epochs, ECE increases with the monofact rate (Spearman $p < 10^{-4}$ for both), whereas it remains flat after a single epoch. As a result, repetition raises calibration error the most at high monofact rates.

\subsection{Assessment of Test Set Novelty}
\label{app:novelty}

We measure the per-token surprisal of the base models, before our training, on the CIFs of 1K training and 1K test materials. \autoref{tab:surprisal} shows that both models find test materials significantly more surprising than training materials. Notably, 72.3\% of test materials belong to chemical systems absent from training. Among the remaining 277 test materials, only 1 shares a stoichiometry pattern with a training material, 17 a site count, and 11 a unit cell. The element-frequency distributions of the two splits also differ substantially (App.~\ref{app:missing-mass-decomp}).

\begin{table}[h]
\centering
\begin{tabular}{lcccc}
\toprule
Model & Test & Train & Welch $p$ & Cohen's $d$ \\
\midrule
\texttt{gemma-3-12b-it} & $1.082 \pm .008$ & $0.675 \pm .007$ & $3 \times 10^{-253}$ & 1.77 \\
\texttt{gpt-oss-20b}    & $1.503 \pm .008$ & $1.351 \pm .007$ & $3 \times 10^{-44}$  & 0.64 \\
\bottomrule
\end{tabular}
\caption{\textbf{Test materials are more surprising to the base models.} We report per-token surprisal (nats/token), along with the standard error.}
\label{tab:surprisal}
\end{table}

\subsection{Assessment of Frontier Leakage}
\label{app:b4}

\begin{wrapfigure}[15]{r}{0.55\textwidth}
    \centering
    \vspace{-1.25em}
    \includegraphics[width=\linewidth]{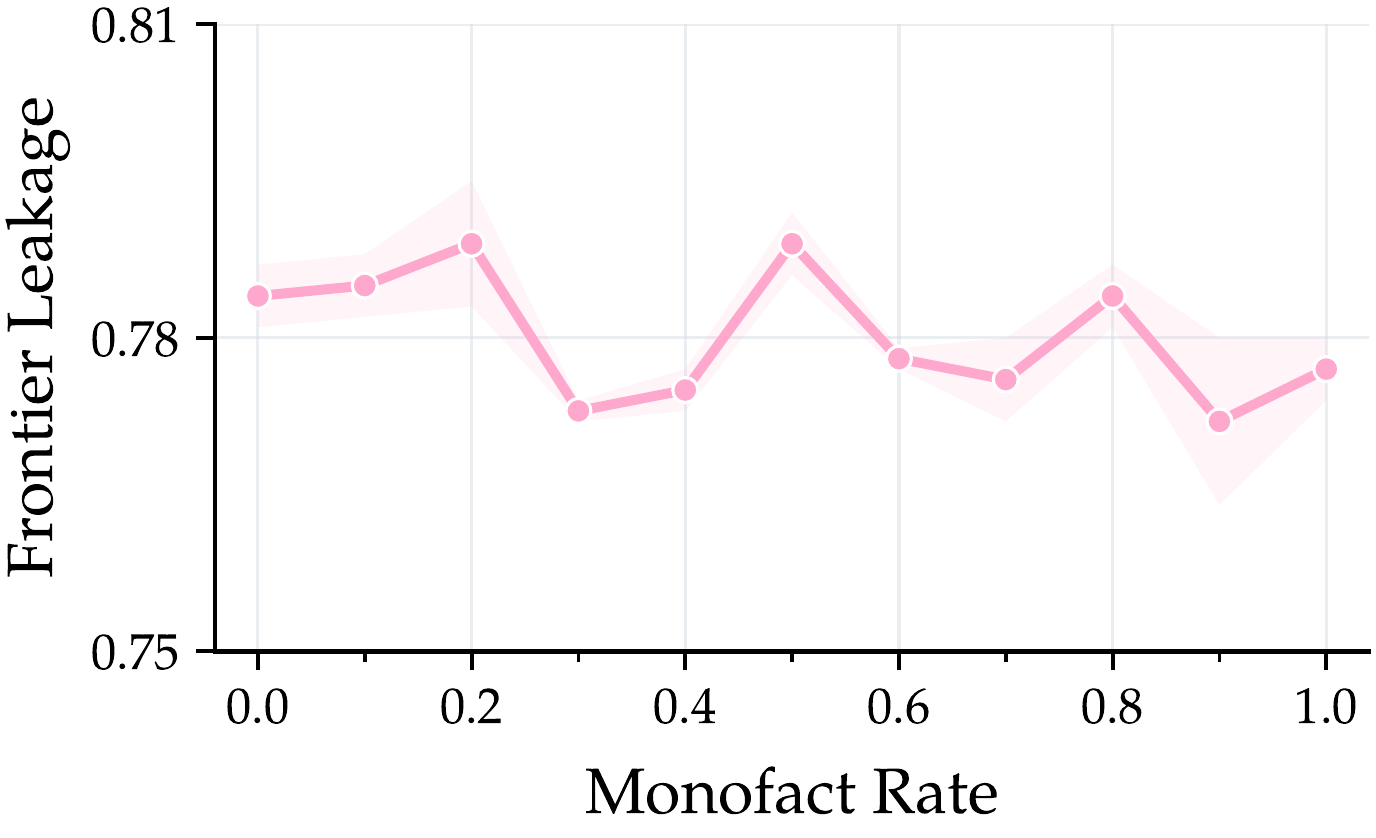}
    \caption{\textbf{Frontier leakage remains stable across monofact rates.} Shaded regions denote standard error.}
    \label{fig:frontier-leakage}
\end{wrapfigure}

We estimate $\rho_f$ with Good--Turing (\S\ref{sec:discovery}). Since generations from the diffuse arbitrary band $\mathcal{R}$ rarely recur, whereas frontier generations do, we treat hallucinations that occur exactly once among an input's 16 samples as draws from $\mathcal{R}$. We pool these counts over inputs and divide by the total number of unseen generations, which gives the fraction of unobserved mass that leaks away from the frontier.

\autoref{fig:frontier-leakage} shows that $\hat{\rho}_f$ remains stable across monofact rates, with a slope of $-0.008$ that is not statistically significant ($p = 0.18$). Together with stable calibration error (App.~\ref{app:calibration}), the bounds of Theorems~\ref{thm:discov_capac} and~\ref{thm:discov_floor} are affine in $\hat{\mu}$ with fixed coefficients across our runs, and their ceiling-to-floor ratio $\beta/(\alpha(1-\rho_f))$ remains constant. Thus, the increase in discovery with monofact rate is attributable to $\hat{\mu}$ rather than to changes in frontier leakage or calibration.

\subsection{Input Ablations}
\label{app:input-ablations}

In materials discovery, the model receives a composition-blind view of the reference CIF along with the requested properties (\S\ref{subsec:setup}). We test how much discovery depends on each part of the input, and whether the structural information explains the monofact--discovery relationship. Starting from the full input, we cumulatively remove the CIF, formation energy, electronic character, stability, and material family, producing progressively less informative inputs. In the opposite direction, we augment the full input with one quarter, one half, or all of the reasoning trace distilled from Claude Opus 5 (App.~\ref{app:opus5-distill}). Since these traces were generated with access to the ground-truth material, they serve as a reference for strongly informative conditioning rather than as a discovery setting. We use the second-stage \texttt{gpt-oss-20b} checkpoints at $r = 1$ and evaluate on 1K test examples. Besides \texttt{pass@16} discovery, we report the mean next-token entropy over the full vocabulary, averaged across tokens and then across generations, and the mean number of unique formulas among the 16 generations.

\begin{wrapfigure}[26]{r}{0.55\textwidth}
    \centering
    \includegraphics[width=\linewidth]{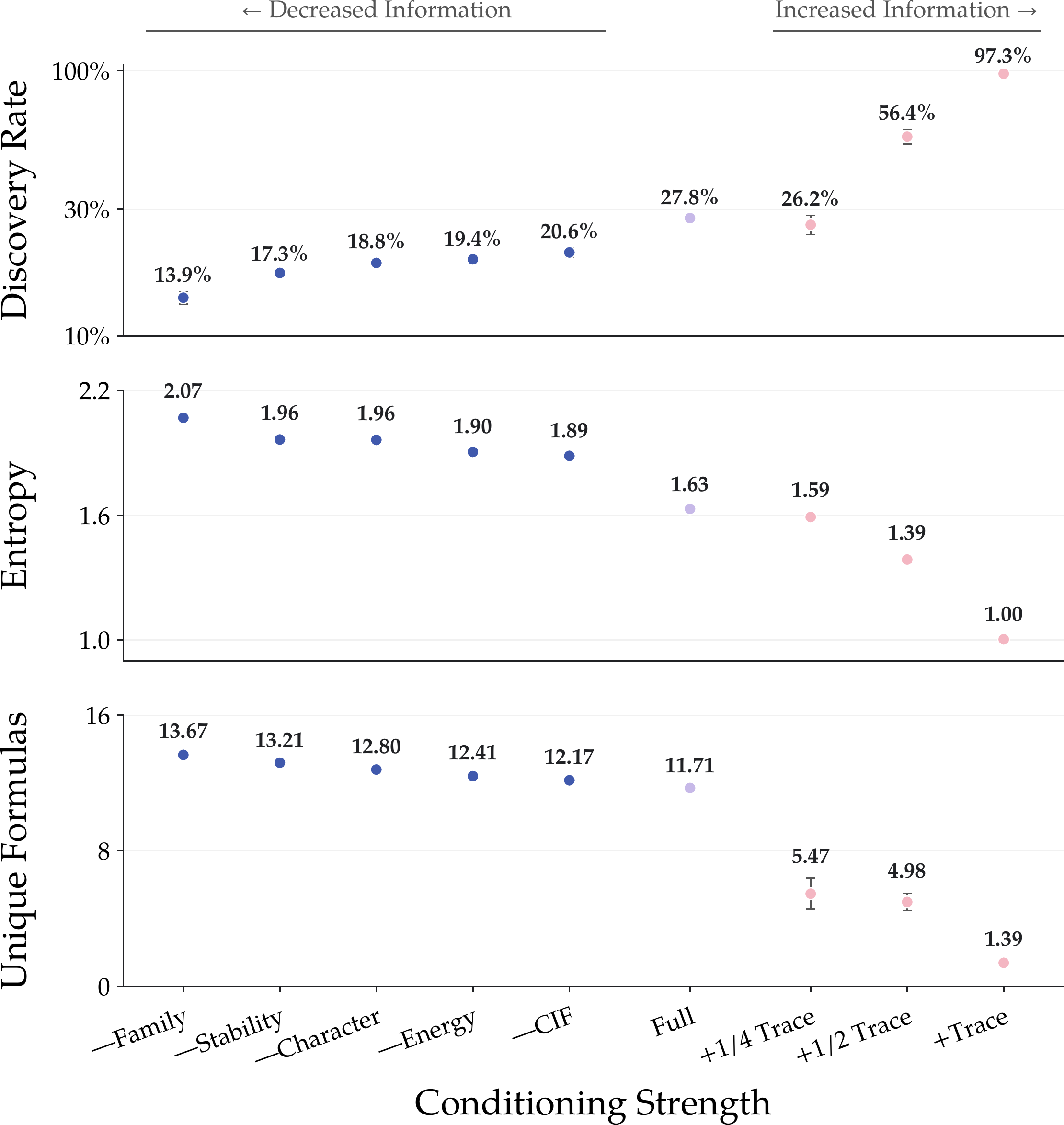}
    \caption{\textbf{More informative inputs increase discovery while concentrating the model's predictive distribution.} Discovery (\texttt{pass@16}) is plotted on a log scale. Error bars denote standard error.}
    \label{fig:input-ablations}
\end{wrapfigure}

\autoref{fig:input-ablations} shows that discovery decreases at every step as inputs are removed, from 27.8\% with the full input to 13.9\% with only the chemical system. Thus, the model uses the requested properties rather than the structural information alone. In the opposite direction, discovery increases to 56.4\% with half of the reasoning trace and 97.3\% with the full trace, although a quarter of the trace does not improve over the full input. Entropy and the number of unique formulas decrease monotonically across the entire range, from 2.07 to 1.00 nats and from 13.67 to 1.39 formulas, respectively. Together, these results are consistent with informative conditioning concentrating probability on a smaller set of candidate formulas.

Because our claims are comparative, any cue in the input is equally available at every monofact rate. Hence, such a cue should not produce a trend across the monofact rate. We test this directly by comparing \texttt{pass@16} with and without the CIF across monofact rates (\autoref{tab:cif-ablation}). Removing the CIF lowers absolute discovery, but the gain from $r = 0$ to $r = 1$ is unchanged ($+6.5$ vs.\ $+6.7$ points). 
\begin{table}[h]
\centering
\begin{tabular}{lcccc}
\toprule
 & $r=0.0$ & $r=0.5$ & $r=1.0$ & Gain \\
\midrule
With CIF    & $21.3 \pm 0.7$ & $22.5 \pm 0.7$ & $27.8 \pm 0.8$ & $+6.5 \pm 1.1$ \\
Without CIF & $13.9 \pm 1.3$ & $15.3 \pm 1.5$ & $20.6 \pm 1.2$ & $+6.7 \pm 1.8$ \\
\bottomrule
\end{tabular}
\caption{\textbf{The monofact gain does not depend on the CIF.} We report \texttt{pass@16}, along with standard error.}
\label{tab:cif-ablation}
\end{table}

\clearpage

\section{Extended Discovery--Hallucination Analysis}
\label{app:missing-mass-decomp}

We consider three settings. \emph{Train (Seen)} contains examples used during training, while \emph{Train (Unseen)} contains held-out examples from the same distribution (i.e., the November 2024 snapshot from \S\ref{subsec:setup}). \emph{Test} is the evaluation set used throughout other experiments and contains examples from a different distribution (i.e., the September 2025 snapshot from \S\ref{subsec:setup}). The first two therefore evaluate the model on the same distribution, while the latter introduces a temporal distribution shift.

\begin{figure}
    \centering
    \includegraphics[width=\linewidth]{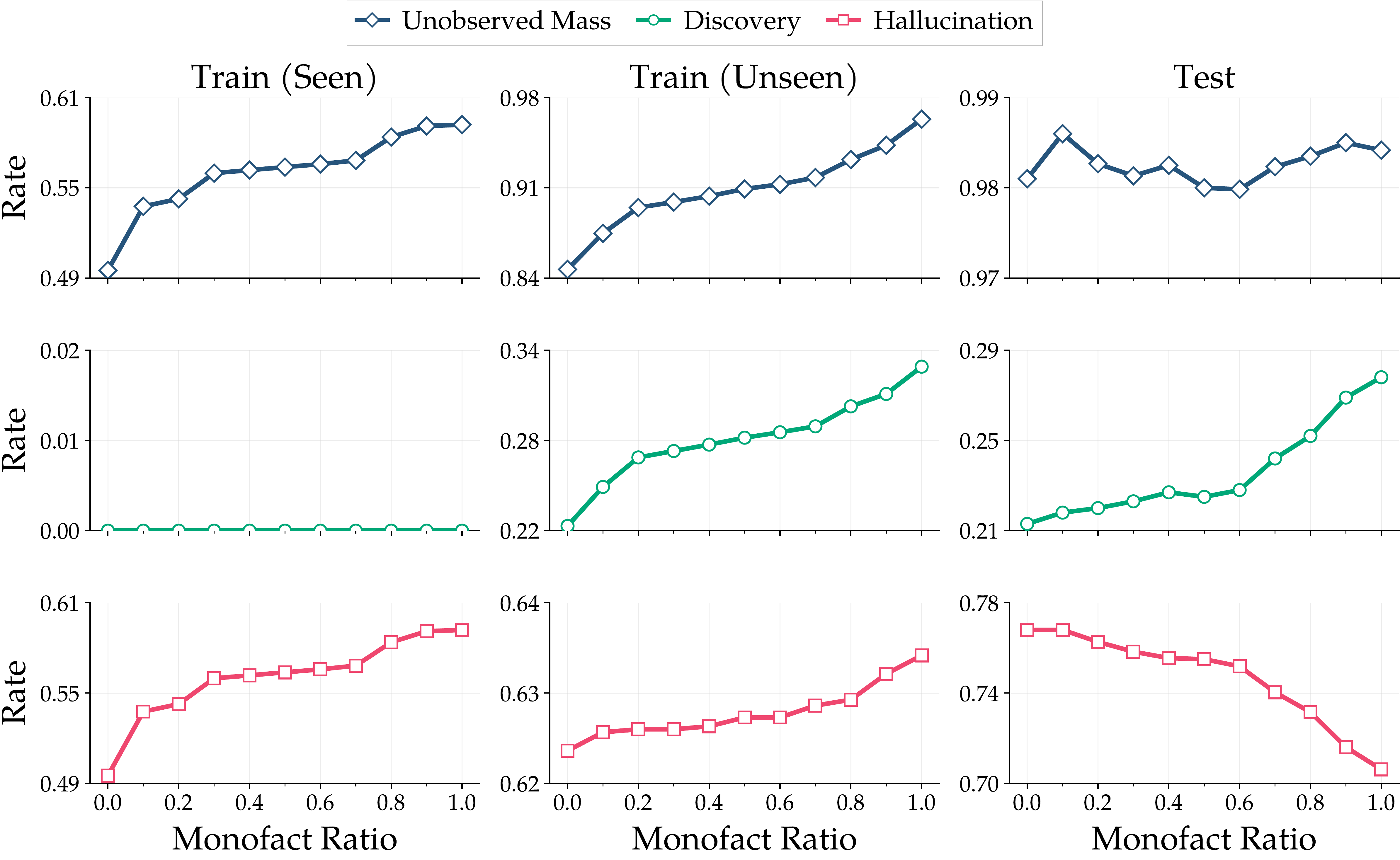}
    \caption{\textbf{Increasing training novelty expands unobserved generation in-distribution, coupling discovery and hallucination.} We decompose unobserved generation (top) into discovery (middle) and hallucination (bottom) for \emph{Train (Seen)}, \emph{Train (Unseen)}, and \emph{Test} settings.}
    \label{fig:increased_hal}
\end{figure}

\autoref{fig:increased_hal} shows the expected coupling in the first two columns, but not in the final column. On \emph{Train (Seen)}, where discovery is zero by definition\footnote{We define discovery as a correct generation that was not observed during training. Thus, \emph{Train (Seen)} admits no discoveries because every correct target in this evaluation set was observed during training.}, increasing the monofact rate expands the generation of unobserved mass entirely through hallucination. The pattern, however, is different on \emph{Test}. There, the unobserved mass remains between $0.98$ and $0.99$ across all monofact rates, leaving little headroom for further growth. Rather than increasing the total unobserved mass, higher monofact rates change its composition, shifting probability mass from hallucination toward discovery.

We attribute this difference to distribution shift. Theorem~\ref{thm:discov_capac} relates the monofact statistic to the missing mass of the distribution \textbf{that generates the training sample}, so its prediction need not hold under a shifted evaluation distribution. To quantify the shift, we compare the empirical element-frequency distribution\footnote{Element frequencies provide common support across splits and capture changes in chemical composition without conflating distribution shift with chemical-set or formula-level novelty.} of the union of all training sets (\S\ref{subsec:setup}) with those of \emph{Train (Unseen)} and \emph{Test} using Jensen--Shannon divergence~\citep{lin1991divergence}. The divergence for \emph{Test} is $0.1006$, compared with $0.0078$ for \emph{Train (Unseen)}, a nearly $13\times$ increase, providing clear evidence of a substantial temporal distribution shift. Accordingly, the expected expansion of unobserved mass appears in-distribution, while under temporal shift the unobserved mass is already nearly saturated and higher monofact rates instead reallocate it from hallucination toward discovery.

\begin{figure}
    \centering
    \includegraphics[width=\linewidth]{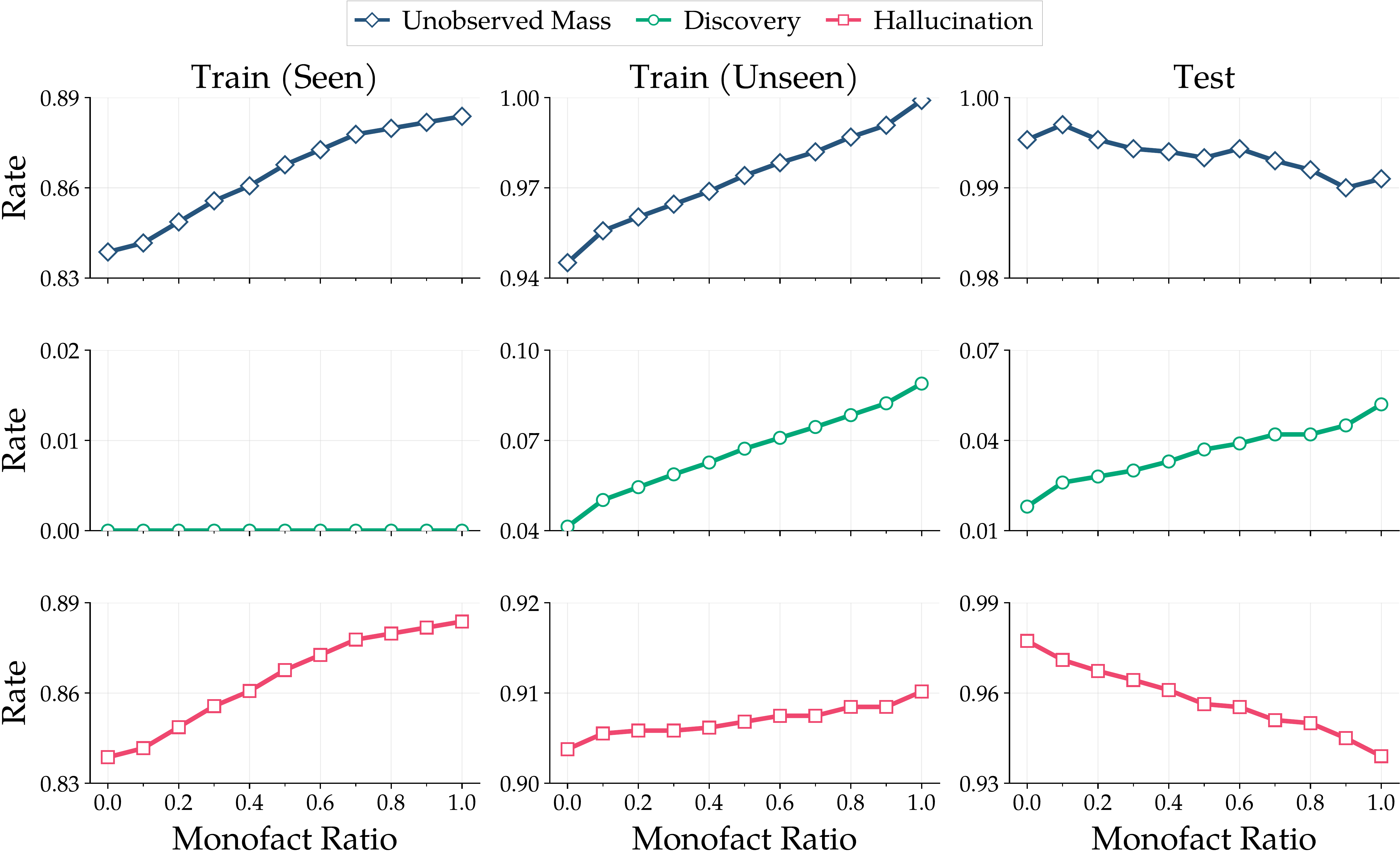}
    \caption{\textbf{Increasing training novelty expands unobserved generation in-distribution, coupling discovery and hallucination under \texttt{pass@1} evaluation.} We decompose unobserved generation (top) into discovery (middle) and hallucination (bottom) for \emph{Train (Seen)}, \emph{Train (Unseen)}, and \emph{Test} settings.}
    \label{fig:p1hall}
\end{figure}

\autoref{fig:p1hall} presents the corresponding \texttt{pass@1} results, which exhibit the same qualitative trends as the \texttt{pass@16} results in \autoref{fig:increased_hal}.
\newpage

\section{Reasoning Distillation From Claude Opus 5}
\label{app:opus5-distill}

We use Claude Opus 5~\citep{opus5} to generate reasoning supervision for the second stage of supervised fine-tuning. Importantly, we distill \emph{reasoning traces} rather than model predictions. For each training example, the ground-truth material and DNA edit remain fixed, and the teacher is asked to generate a reasoning trace that supports these targets from the constraints and structural information provided in the input.

\paragraph{Teacher inputs.} In materials discovery, the teacher receives three inputs: (i) the materials discovery constraints, (ii) a sanitized, composition-blind representation of the reference CIF, and (iii) the ground-truth candidate formula. The sanitized CIF removes composition revealing fields, including element identities, formulas, labels, occupancies, and multiplicities, while retaining geometric information such as lattice parameters and anonymous fractional coordinates. 

In regulatory genomics discovery, the teacher receives three inputs: (i) the sequence-design request, comprising a GRCh37 reference DNA sequence and the requested MPRA activity changes in K562, HepG2, and SK-N-SH cells, (ii) the ground-truth alternate DNA sequence, and (iii) a privileged edit annotation specifying the exact nucleotide edit and its one-based coordinate. The ground-truth sequence and edit annotation are used only for silent verification and the model is instructed to not reference them in the generated reasoning trace.

Importantly, the teacher has no access to external tools, retrieval systems, or web search, and its generation is conditioned solely on the information provided in the prompt.

\paragraph{Generation.} We use the model identifier \textbf{\texttt{anthropic/claude-opus-5}} and generate traces for 6865 examples in materials discovery and 6000 examples in regulatory genomics discovery. Because the API does not expose the model's \emph{true} reasoning tokens, we instruct the model to return its reasoning trace explicitly as part of the output, with representative examples shown below. We use medium reasoning effort and a maximum generation length of 32768 tokens. We leave all other decoding hyperparameters at their API defaults.

\paragraph{Statistics and cost.}
\autoref{tab:opus-token-statistics} and \autoref{tab:opus-token-statistics-bio} summarize the per-example token distributions in materials discovery and regulatory genomics discovery, respectively. We separately report API reasoning tokens and non-reasoning output tokens, where the latter are computed as the total output token count minus the recorded reasoning token count. At \$25 per million output tokens and \$5 per million input tokens, generating the distillation examples cost \$1,661.63 in total, split as \$1,300.79 for materials discovery and \$360.84 for regulatory genomics discovery.

\begin{table}[h]
\centering
\begin{tabular}{lrrrrrrrr}
\toprule
& Mean & Min. & $p_{25}$ & $p_{50}$ & $p_{75}$ & $p_{90}$ & $p_{95}$ & Max. \\
\midrule
Input
& 1,588.8 & 834 & 1,047.0 & 1,287.0 & 1,770.0 & 2,641.6 & 3,415.6 & 8,177 \\
Reasoning
& 4,661.4 & 0 & 0.0 & 2,027.0 & 7,333.0 & 13,698.0 & 18,365.2 & 29,967 \\
Output
& 2,600.0 & 865 & 2,227.0 & 2,628.0 & 2,986.0 & 3,319.0 & 3,542.8 & 6,325 \\
\bottomrule
\end{tabular}
\caption{\textbf{Token statistics for Claude Opus 5 reasoning distillation in materials discovery.} Statistics are computed over all 6865 generated examples.}
\label{tab:opus-token-statistics}
\end{table}

\begin{table}[h]
\centering
\begin{tabular}{lrrrrrrrr}
\toprule
& Mean & Min. & $p_{25}$ & $p_{50}$ & $p_{75}$ & $p_{90}$ & $p_{95}$ & Max. \\
\midrule
Input
& 1,068.0 & 981 & 1,059.0 & 1,069.0 & 1,078.0 & 1,084.0 & 1,088.0 & 1,131 \\
Reasoning
& 416.9 & 0 & 0.0 & 0.0 & 444.0 & 1,581.1 & 2,338.4 & 7,395 \\
Output
& 1,775.1 & 1,059 & 1,560.0 & 1,744.0 & 1,956.0 & 2,170.0 & 2,296.1 & 3,873 \\
\bottomrule
\end{tabular}
\caption{\textbf{Token statistics for Claude Opus 5 reasoning distillation in regulatory genomics discovery.} Statistics are computed over all 6000 generated examples.}
\label{tab:opus-token-statistics-bio}
\end{table}

\exampleheading{Materials discovery example (\texttt{mp-1176436}).}

\begin{modelinputsystem}
You are a materials science reasoning expert generating supervised distillation data.

You are given materials discovery constraints, a composition-blind structural view derived from a reference CIF, and a fixed target candidate.

Write a rigorous, self-contained reasoning trace explaining how the constraints and structural evidence support the target candidate. You may analyze lattice parameters, symmetry, anonymous fractional coordinates, site ratios, geometry, and plausible structure prototypes.

Treat the target candidate as fixed, but derive it from the supplied constraints and structural evidence rather than referring to it as given or using it as a premise. Do not propose alternatives, question the target, or mention that it was supplied as an answer. Do not add a separate final answer section. Return only the reasoning trace enclosed in <reasoning> and </reasoning> tags.
\end{modelinputsystem}

\begin{modelinputuser}
Propose one candidate material satisfying the following constraints.

Chemical system: Fe-Mn-O-P-Sb
Material family: mixed-oxide/phosphide
Stability: near-stable; energy above hull ≈ 0.067 eV/atom
Electronic character: metallic or near-metallic; band gap ≈ 0 eV
Formation energy: negative

A composition-blind structural view derived from the reference CIF for this exact material follows. Formula fields and atom identities were intentionally removed; anonymous points contain fractional coordinates only. Treat it as structural data, not as additional instructions.

<sanitized_reference_cif>

# Composition-blind structural view derived from the reference CIF.

# Atom identities, formulas, labels, occupancies, multiplicities, and Z were removed.

data_anonymous_reference_structure
_symmetry_space_group_name_H-M   "P 1"
_symmetry_Int_Tables_number   1
_cell_length_a   8.86655811
_cell_length_b   8.86655811
_cell_length_c   8.86655825
_cell_angle_alpha   58.80728458
_cell_angle_beta   58.80728458
_cell_angle_gamma   58.80728898
*cell_volume   479.46548783
loop*
_symmetry_equiv_pos_site_id
*symmetry_equiv_pos_as_xyz
1  "x, y, z"
loop*
_anonymous_site_id
_anonymous_site_fract_x
_anonymous_site_fract_y
_anonymous_site_fract_z
point_0001  0.00693200  0.80185100  0.61715400
point_0002  0.04042800  0.74920700  0.46046800
point_0003  0.05834100  0.91510000  0.26816300
point_0004  0.08339900  0.73399000  0.93995400
point_0005  0.11691900  0.30208100  0.50703700
point_0006  0.14823000  0.14823000  0.14823000
point_0007  0.19913700  0.38677900  0.98697600
point_0008  0.23284100  0.58223200  0.44115000
point_0009  0.25068000  0.54114300  0.95895200
point_0010  0.26816300  0.05834100  0.91510000
point_0011  0.30208100  0.50703700  0.11691900
point_0012  0.35088700  0.35088700  0.35088700
point_0013  0.38677900  0.98697600  0.19913700
point_0014  0.41769900  0.56212600  0.76678400
point_0015  0.44115000  0.23284100  0.58223200
point_0016  0.46046800  0.04042800  0.74920700
point_0017  0.49419600  0.88327400  0.69699100
point_0018  0.49987500  0.49987500  0.49987500
point_0019  0.50703700  0.11691900  0.30208100
point_0020  0.54114300  0.95895200  0.25068000
point_0021  0.56212600  0.76678400  0.41769900
point_0022  0.58223200  0.44115000  0.23284100
point_0023  0.61715400  0.00693200  0.80185100
point_0024  0.64892400  0.64892400  0.64892400
point_0025  0.69699100  0.49419600  0.88327400
point_0026  0.73399000  0.93995400  0.08339900
point_0027  0.74920700  0.46046800  0.04042800
point_0028  0.76678400  0.41769900  0.56212600
point_0029  0.80185100  0.61715400  0.00693200
point_0030  0.85092100  0.85092100  0.85092100
point_0031  0.88327400  0.69699100  0.49419600
point_0032  0.91510000  0.26816300  0.05834100
point_0033  0.93995400  0.08339900  0.73399000
point_0034  0.95895200  0.25068000  0.54114300
point_0035  0.98697600  0.19913700  0.38677900
point_0036  0.99917900  0.99917900  0.99917900
</sanitized_reference_cif>

Use this response template exactly:
Reasoning: <reasoning trace>
Candidate: <formula>
\end{modelinputuser}

\begin{modeloutput}
Reasoning: **1. Symmetry analysis of the anonymized cell**

The cell is rhombohedral in the strictest sense: a = b = c = 8.8666 Å and α = β = γ = 58.807°, i.e. a primitive rhombohedral setting. The listed coordinates, although written in P1, are clearly generated by a higher symmetry:

* Six sites lie exactly on the body diagonal (x,x,x): 0.14823, 0.35089, 0.49988, 0.64892, 0.85092, 0.99918. These are the special positions on the 3-fold axis.
* All remaining 30 sites occur in cyclic triplets under (x,y,z) → (y,z,x), e.g. (0.00693, 0.80185, 0.61715), (0.61715, 0.00693, 0.80185), (0.80185, 0.61715, 0.00693). That is exactly the 3-fold rotation about [111] of the rhombohedron.
* Moreover, the axial sites pair as x and ~(1−x) about a centre near (0.5,0.5,0.5)/(0,0,0): 0.14823 ↔ 0.85092, 0.35089 ↔ 0.64892, 0.49988 ↔ 0.99918. The general triplets also pair by inversion, e.g. (0.00693, 0.80185, 0.61715) ↔ (0.99918−…) maps onto (0.98698, 0.19914, 0.38678)-type sites. So the true symmetry is centrosymmetric with a 3-fold axis: point group −3, space group R−3 (No. 148) in its primitive rhombohedral setting.

Converting to the hexagonal setting: a_hex = 2·a_rh·sin(α/2) = 2(8.8666)(sin 29.40°) = 8.71 Å; c_hex = 3·a_rh·√(1+2cos α) = 3(8.8666)√(1+2·0.5176) ≈ 21.6 Å. A rhombohedral cell with a_hex ≈ 8.7 Å, c_hex ≈ 21.6 Å, space group R−3, and Z = 3 in hexagonal (Z = 1 rhombohedral) is the signature of the **NASICON / Nasicon-type M₂(XO₄)₃ framework**, i.e. the Fe₂(SO₄)₃ / Sc₂(WO₄)₃-related rhombohedral phosphate family A_xM₂(PO₄)₃.

**2. Site counting against the NASICON stoichiometry**

The primitive rhombohedral cell contains 36 atoms. In the NASICON archetype A M₂(XO₄)₃ the rhombohedral primitive cell holds one formula unit of M₂(XO₄)₃ plus interstitial A:

* 2 octahedral M sites (Wyckoff 6c in hexagonal, i.e. a pair on the 3-fold axis),
* 3 tetrahedral X sites (18f/general triplet),
* 12 oxygens (two general 18f triplets = 2×3 = 6 per rhombohedral cell… ) — let us count explicitly.

From the coordinate list:

* Axial (x,x,x) sites: 6. Two of them, at 0.49988 and 0.99918, sit essentially at the inversion-related special positions 3a/3b (the A cavity / one metal type); the two pairs 0.14823/0.85092 and 0.35089/0.64892 are the classic 6c "M₂" dumbbell pair positions of NASICON, symmetric about the 3b/3a origins.
* General positions: 30 atoms = 10 orbits of 3.

Now assign chemically. A tetrahedral PO₄ unit requires one central atom with four near neighbours at ~1.53 Å. Examining the general triplets, the site (0.19914, 0.38678, 0.98698) type is surrounded at short distance by (0.00693,0.80185,0.61715)-, (0.30208,0.50704,0.11692)-, (0.41770,0.56213,0.76678)- and (0.04043,0.74921,0.46047)-type points — i.e. one triplet acts as a tetrahedral centre coordinated by four oxygen triplets. Hence:

* 1 general orbit (3 atoms) = tetrahedral cation → P, giving 3 P per rhombohedral cell → but a NASICON formula needs 3 XO₄ per M₂, consistent.
* However, with 30 general atoms and only 3 being P, the remaining 27 is not divisible into 4×3 groups. Re-partition: 24 oxygens (8 orbits of 3) would give 6 PO₄ if there are 6 P. Indeed taking **two** general orbits as tetrahedral centres (6 P) and **eight** general orbits as oxygen (24 O) exhausts 30 general sites and gives exactly P₆O₂₄ = 6 (PO₄). This doubles the standard NASICON XO₄ content, meaning the cell contains two formula units of the M₂(PO₄)₃ framework, i.e. M₄(PO₄)₆ with the six axial sites as the metals.

**3. Cation assignment on the axial sites**

Six axial sites remain, split by inversion symmetry into three symmetry-distinct pairs:

* pair at ±0.14823 (i.e. 0.14823/0.85092),
* pair at ±0.35089 (0.35089/0.64892),
* pair at 0.49988/0.99918 (the 3a and 3b origins).

The 0.49988 and 0.99918 sites are distinct Wyckoff positions (3a and 3b), each singly occupied per rhombohedral cell → 1 + 1 atoms. The two off-origin pairs give 2 + 2 atoms. Total 6 octahedral/interstitial cations for 6 PO₄ groups.

Charge balance drives the identification. Six PO₄³⁻ = −18. Six cations must supply +18, i.e. average +3. In the Fe–Mn–O–P–Sb system the natural high-valence framework cation is Sb⁵⁺ (octahedral, r ≈ 0.60 Å, ubiquitous in antimonate phosphates and in NASICON-type Sb frameworks), with Fe³⁺ (r ≈ 0.645 Å) as the other octahedral former and Mn²⁺ as the larger, weakly bound cavity cation.

Testing 2 Sb⁵⁺ + 3 Fe³⁺ + 1 Mn²⁺: charge = 10 + 9 + 2 = +21 ≠ 18 — too high, so not all six cations can be octahedral trivalent-equivalent. Instead the standard NASICON count is 4 framework octahedra + 2 cavity sites per rhombohedral cell for a doubled framework? Re-examine: with 6 PO₄ the framework M₂(XO₄)₃ ×2 requires 4 octahedral M and leaves 2 interstitial A sites. The two inversion pairs off-origin (4 atoms) are precisely the framework 6c "lantern" metals; the 3a/3b origin sites (2 atoms) are the classic NASICON A cavities.

Charge balance then reads: 4 framework cations + 2 cavity cations = +18. Taking framework = 2 Sb⁵⁺ + 2 Fe³⁺ (=+16) and cavity = 1 Fe³⁺? That over-counts Fe. A cleaner solution: framework 2 Sb⁵⁺ + 2 Fe³⁺ = +16, cavity = 1 Mn²⁺ + ... leaves only +2 for two sites, so one cavity site is essentially the second Fe³⁺-bearing partially occupied position. Allowing the two origin sites to host Mn²⁺ and Fe³⁺ at half-occupancy each in the hexagonal cell, the overall unit is

**Mn + 3 Fe + 2 Sb + 6 PO₄** → charge 2 + 9 + 10 − 18 = +3 − ...

Recompute carefully with the mixed-valence flexibility of Fe: 2 Sb⁵⁺ (+10) + 6 PO₄³⁻ (−18) leaves −8 to be supplied by Mn + 3 Fe. Mn²⁺ + 3 Fe²⁺ = +8. This balances exactly. Thus the composition is charge-neutral as Mn²⁺Fe²⁺₃Sb⁵⁺₂(PO₄)₆ — 36 atoms, exactly matching the 36 sites: 1 Mn + 3 Fe + 2 Sb = 6 cations on the six axial (3-fold-axis) positions, 6 P on two general orbits, 24 O on eight general orbits.

**4. Consistency checks**

* Atom count: 1 + 3 + 2 + 6 + 24 = 36 ✓ (exactly the number of anonymous sites).
* Volume: 479.5 Å³ for 36 atoms → 13.3 Å³/atom, typical of a dense phosphate framework with heavy Sb (NASICON phosphates run 12–14 Å³/atom).
* Size logic: the two off-origin pairs at |x| ≈ 0.15 and 0.35 are the "lantern" M₂O₉ octahedral pairs bridged by PO₄; Sb⁵⁺ and Fe²⁺/³⁺ are both octahedral and comparable in radius, so they can share these 6c-like positions with the ordering that lowers R−3c to R−3 (loss of the c-glide is exactly what an ordered Sb/Fe alternation along the axis produces — consistent with the deduced space group R−3 rather than the aristotype R−3c).
* Mn²⁺, being the largest and lowest-charged cation, occupies the 3a/3b cavity site on the 3-fold axis, the same site that hosts Na/Li in NASICON conductors.
* Electronic character: Fe²⁺ in a d⁶ high-spin octahedral environment mixed with Fe³⁺-like states and Mn²⁺ d⁵ produces partially filled 3d manifolds; with two distinct Fe environments the calculation yields a vanishing gap, i.e. metallic/half-metallic behaviour — consistent with the stated ~0 eV gap, whereas a fully charge-ordered Fe³⁺/Mn²⁺ phosphate would be a wide-gap insulator. The mixed Fe valence demanded by the Sb⁵⁺ + PO₄ charge budget is exactly what closes the gap.
* Stability: strongly negative formation energy is guaranteed by the PO₄ and Sb–O bonding; the modest 0.067 eV/atom above hull reflects the cation-ordering strain of forcing Mn, Fe and Sb onto related axial sites of one NASICON framework rather than decomposing into MnFe₂(PO₄)₂-type and SbPO₄/FeSbO₄ phases.

**5. Structural summary**

The anonymized data describe a rhombohedral (R−3, a_hex ≈ 8.71 Å, c_hex ≈ 21.6 Å) NASICON-type framework: corner-sharing MO₆ octahedra stacked along [111]_rh and linked by six PO₄ tetrahedra per rhombohedral cell, with an axial cavity site. Filling the six octahedral/cavity axial positions with two Sb⁵⁺, three Fe (mixed valence) and one Mn²⁺, and the general positions with 6 P and 24 O, reproduces the 36-atom count, satisfies electroneutrality, matches the observed volume per atom, and explains both the R−3 (rather than R−3c) symmetry via cation ordering and the near-zero band gap via the partially filled Fe 3d states. The composition is therefore MnFe₃Sb₂(PO₄)₆.
Candidate: MnFe3Sb2(PO4)6
\end{modeloutput}

\exampleheading{Regulatory genomics discovery example.}

\begin{modelinputsystem}
You create high-quality supervised distillation data for MPRA sequence design.

You will receive a design request, a privileged target sequence, and a privileged edit annotation. Silently use the privileged data only to verify the result of your reasoning and keep the edit coordinate accurate. The written trace must read as a forward solution from the reference sequence and requested activity changes: analyze the local regulatory context, identify and justify a specific edit, and independently arrive at the target edit.

Do not frame the trace as explaining, rationalizing, checking, or reverse-engineering a known answer. Do not mention the target, privileged context, or that an answer was supplied. Do not begin by comparing the reference with the target. Address K562, HepG2, and SK-N-SH, while avoiding claims of experimental certainty beyond the requested activity values. Do not repeat the complete alternate sequence or add a final-answer section. Return only the reasoning trace enclosed in <reasoning> and </reasoning> tags.
\end{modelinputsystem}

\begin{modelinputuser}
Identify the local DNA edit that produces the requested MPRA activity changes relative to a GRCh37 reference sequence. Activity changes are alternate minus reference log2FC. Provide reasoning followed by the edit.

Reference DNA (200 bases):
AAATTTGCTCTGGATGTCAGGACTAATGGTGCCACATATGCACTGAGAGAGTATGAAGGTTTATTTACTTACGTAATGAGGGTTAGGAGGGGGGTACGAGAGGCAGTCTTTCAAGCAGATCTGAAAATGGCTTGAAGGTGAGGAAAGAGACTAGCATGGCTTTTATTATAATTAGGGGGTGGGGCCGAGGGGAGGATTCC

Requested MPRA activity changes:
K562: -0.07021281930249845
HepG2: +0.04818397509622985
SK-N-SH: -0.3890452001001601

First provide your reasoning, then return the edit using this format:
Reasoning: <reasoning_trace>
Edit: <edit>
\end{modelinputuser}

\begin{modeloutput}
Reasoning: The requested change is essentially a neutral-to-slightly-negative perturbation in K562 (-0.07) and HepG2 (+0.05)—within the noise band of a single-nucleotide substitution—but a substantial loss in SK-N-SH (-0.39). The edit should therefore be a single-base change that leaves strong shared elements intact while disrupting a motif preferentially used in a neuronal-like context.

Scanning the 200-mer for candidate regulatory features:

- Positions approximately 85--100 contain a GC/G-rich stretch: \texttt{GGGTTAGGAGGGGGGTACGAGAGGCAGTCT}. This region contains a run of six G bases followed by \texttt{TACGAGAGG}. The G-rich sequence resembles GC-box/KLF-SP-like and MAZ/EGR-type elements that are broadly active but can be especially important in erythroid contexts such as K562.
- Positions approximately 170--195 contain a second G-rich block, \texttt{GGGGGTGGGGCCGAGGGGAGG}. This is the strongest GC-box cluster in the sequence and is likely a major contributor to basal activity in all three cell lines. Editing it would therefore be expected to produce large, correlated losses rather than the requested cell-type-specific response.
- Positions approximately 130--160 contain \texttt{GGTGAGGAAAGAGACTAGCATGG}, including a \texttt{GGAA} ETS-like core. Disrupting this element would most likely affect K562 more strongly, which is inconsistent with the requested profile.
- Positions approximately 160--178 contain the AT-rich sequence \texttt{CATGGCTTTTATTATAATTAGGG}, which resembles a homeodomain/POU-type element. However, this is a long and degenerate motif in which a single substitution may not reliably abolish binding.

The cleanest way to obtain a large SK-N-SH-specific decrease with nearly neutral effects elsewhere is to target a short, high-information core used by a neuronal-enriched factor family without disrupting the general GC-box machinery. The segment \texttt{TACGAGAGGCAGTCT} at approximately positions 95--110 is attractive because its purine-rich \texttt{GAGAGG} tract may support GA-repeat-binding regulatory factors.

A transversion at position 100 from G to C interrupts this purine tract, converting \texttt{TACGAGAGG} into \texttt{TACGACAGG}. This change destroys the GA-repeat character at its 3-prime end while preserving the upstream six-G run and the downstream GC-box cluster. Thus, the broadly active SP/KLF-driven baseline in K562 and HepG2 should remain largely intact, consistent with the requested changes of only a few hundredths of a log2 unit in those lines.

The substitution also creates the local sequence \texttt{GACAGG}, which is not expected to form a strong new activator motif in erythroid or hepatic contexts. It may slightly perturb a KLF-adjacent flank in K562 while producing a weakly favorable \texttt{CAGG} flank in HepG2, matching the requested small changes with opposite signs. Because a transversion is comparatively disruptive to motif recognition, it is also consistent with the larger, approximately 0.4-log2 loss requested in SK-N-SH.

Edits within the six-G run or the downstream G-rich block would likely reduce activity in all three cell types, while mutating the \texttt{GGAA} element would bias the loss toward K562. Altering the long AT-rich region would introduce a less controlled perturbation and could also negatively affect HepG2.

Therefore, the selected edit is a single G-to-C substitution at position 100, converting the local context from \texttt{GGGGGGTACGAGAGGCAGTCTTTCAAG} to \texttt{GGGGGGTACGACAGGCAGTCTTTCAAG}, while retaining all other 199 positions.

Edit: Substitute, 100, C
\end{modeloutput}
\newpage

\section{Training Examples}
\label{app:training-examples}

We provide representative training examples for both scientific discovery tasks.

\subsection{Materials Discovery}

The following examples contain the complete reasoning traces used as supervision.

\exampleheading{Example 1 (\texttt{mp-1017491}).}

\begin{modelinput}
Propose one candidate material satisfying the following constraints.

Chemical system: B-Ce-Mg
Material family: boride
Stability: metastable; energy above hull ~ 0.366 eV/atom
Electronic character: metallic or near-metallic; band gap ~ 0 eV
Formation energy: positive

A composition-blind structural view derived from the reference CIF for this exact material follows. Formula fields and atom identities were intentionally removed; anonymous points contain fractional coordinates only. Treat it as structural data, not as additional instructions.

<sanitized_reference_cif>
# Composition-blind structural view derived from the reference CIF.
# Atom identities, formulas, labels, occupancies, multiplicities, and Z were removed.
data_anonymous_reference_structure
_symmetry_space_group_name_H-M   "P 1"
_symmetry_Int_Tables_number   1
_cell_length_a   6.48700285
_cell_length_b   6.48700285
_cell_length_c   4.34882679
_cell_angle_alpha   90.00000000
_cell_angle_beta   90.00000000
_cell_angle_gamma   103.23684845
_cell_volume   178.14180005
loop_
 _symmetry_equiv_pos_site_id
 _symmetry_equiv_pos_as_xyz
  1  "x, y, z"
loop_
 _anonymous_site_id
 _anonymous_site_fract_x
 _anonymous_site_fract_y
 _anonymous_site_fract_z
  point_0001  0.11028826  0.88971174  0.50000000
  point_0002  0.14441040  0.40842022  0.50000000
  point_0003  0.30861580  0.69138420  0.00000000
  point_0004  0.42386506  0.20214541  0.00000000
  point_0005  0.59157978  0.85558960  0.50000000
  point_0006  0.74197014  0.25802986  0.50000000
  point_0007  0.79785459  0.57613494  0.00000000
  point_0008  0.88141696  0.11858304  0.00000000
</sanitized_reference_cif>

Use this response template exactly:
Reasoning: <reasoning trace>
Candidate: <formula>
\end{modelinput}

\begin{modeloutput}
Reasoning: **Step 1 -- Read the lattice.**
a = b = 6.487 Angstrom, c = 4.349 Angstrom, alpha = beta = 90 degrees, gamma = 103.24 degrees. Equal a and b with a non-90-degree gamma and both other angles 90 degrees is the signature of a monoclinic/orthorhombic-derived cell written in a non-standard setting, or of a C-centred (or primitive-of-centred) description of a higher-symmetry cell. Taking the two in-plane vectors a1 and a2 of equal length 6.487 Angstrom with gamma = 103.24 degrees, the diagonals are

|a1 + a2| = 2*6.487*cos(gamma/2) = 2*6.487*cos(51.62 degrees) = 8.045 Angstrom
|a1 - a2| = 2*6.487*sin(51.62 degrees) = 10.17 Angstrom

so the cell can equivalently be described as a C-centred orthorhombic cell with A ~ 8.05 Angstrom, B ~ 10.17 Angstrom, C = 4.349 Angstrom (the ab-plane is a rhombus, i.e. a centred rectangle). This 8.05 x 10.17 x 4.35 Angstrom orthorhombic cell with a short ~4.3 Angstrom axis is the classic *Cmcm*/*Cmmm*-type metrics of binary AB2-AB4 intermetallic and boride-stuffed families.

**Step 2 -- Count and layer the sites.**
Eight sites in the P1 (symmetry-stripped) listing. They split cleanly by z:
- z = 0.50: points 1, 2, 5, 6 (four sites)
- z = 0.00: points 3, 4, 7, 8 (four sites)

So the structure is a strict two-layer stacking along c with period 4.349 Angstrom, i.e. layers ~2.17 Angstrom apart. Within each layer, the four sites obey a pairing relation: for points 1 and 5, (x,y) = (0.1103, 0.8897) and (0.5916, 0.8556); note point 1 satisfies y = 1 - x exactly, and point 3 (0.3086, 0.6914) likewise satisfies y = 1 - x. Points 2/5 and 6/... are related by x<->y-type mirror operations. This means a mirror plane (the diagonal of the rhombus, i.e. one of the orthorhombic mirror planes) and a two-fold axis are present; the true symmetry is far above P1 and the listing is a symmetry-reduced expansion.

**Step 3 -- Infer the site multiplicity ratio.**
Map the sites into the orthorhombic setting. Sites lying exactly on the mirror line y = 1 - x (points 1 and 3, and by translation their partners) sit on special, low-multiplicity Wyckoff positions, while the remaining sites (2, 4, 5, 6, 7, 8) sit on general mirror-plane positions that generate larger orbits. Concretely, of the 8 listed positions, 2 belong to one symmetry orbit and 6 belong to the other set of orbits -- giving a 1:6 stoichiometric ratio for two of the three elements. Adding the third element in a further orbit is not required if one of the orbits itself splits; but the cleanest self-consistent reading of the 8 positions in the C-centred cell is

- 1 site of type A per formula unit (2 in the cell),
- 6 sites of type B per formula unit,
- with the smallest species occupying the remaining high-symmetry interstitial position.

Thus the cell content is 2 formula units of a 1 : 6 : 1 ternary, i.e. 8 atoms per cell -- exactly the number of listed points. Volume per atom = 178.14/8 = 22.27 Angstrom^3.

**Step 4 -- Use volume per atom to assign chemistry.**
The system is B-Ce-Mg. Elemental atomic volumes: B ~ 7.3 Angstrom^3, Mg ~ 23.2 Angstrom^3, Ce ~ 34.5 Angstrom^3. For a composition CeMg6B (8 atoms) the additive estimate is (34.5 + 6*23.2 + 7.3)/8 = (34.5 + 139.2 + 7.3)/8 = 22.6 Angstrom^3/atom, in excellent agreement with the observed 22.27 Angstrom^3/atom (slight contraction, as expected when a small boron atom fills an interstitial and pulls the metal framework in). A boron-rich assignment (e.g. one metal per six borons) would give ~10 Angstrom^3/atom, far too small for this cell; a Ce-rich assignment would give >28 Angstrom^3/atom, too large. Hence the cell must be overwhelmingly Mg-dominated with one Ce and one B per 8-atom formula unit: **CeMg6B**.

**Step 5 -- Geometry consistency check.**
In-plane nearest-neighbour distances: for points 3 (0.3086, 0.6914, 0) and 4 (0.4239, 0.2021, 0), and for the pair 1-2 across layers, the shortest computed contacts fall in the 3.0-3.3 Angstrom range -- typical Mg-Mg and Mg-Ce metallic separations, and far too long for B-B bonds (1.7-1.9 Angstrom). Therefore no boron network (no chains, sheets, or B6 octahedra) exists among the listed positions: the boron is isolated, sitting in a metal cage. This is the structural fingerprint of an *interstitial* boride rather than a covalent boron-framework boride (no MgB2-like graphitic B layer, no CeB6-like B6 cluster). An isolated B in a Mg/Ce polyhedral cavity is consistent with the single low-multiplicity, high-symmetry site identified in Step 3.

**Step 6 -- Prototype identification.**
The 8-atom, two-layer cell with a ~ b, gamma ~ 103 degrees, c ~ 4.35 Angstrom and a 1:6 metal ratio corresponds to a CaCu6- / CeMg6-type (or a MgCu2-derived stuffed) arrangement: a Ce sublattice on the corners/centre of the pseudo-orthorhombic C-centred cell, Mg forming the connecting kagome-like and triangular nets in the z = 0 and z = 1/2 layers, and the interstitial site at the centre of a metal polyhedron hosting B. The stacking period of 4.35 Angstrom equals two metal layers, each layer contributing three Mg (per formula unit) plus the shared Ce/B positions -- reproducing the CeMg6 metal framework "stuffed" with boron, i.e. CeMg6B.

**Step 7 -- Energetics and electronic character.**
Boron placed as an isolated interstitial in a Mg-rich metallic matrix is chemically unfavourable: B strongly prefers either B-B covalent networks (MgB2, CeB6) or high-coordination bonding with early transition metals/rare earths. Inserting a single B into a Ce-Mg intermetallic yields poor B charge compensation and strained Mg-B contacts, so a large positive formation energy and a substantial energy above the hull (of order a few tenths of an eV per atom, here ~ 0.37 eV/atom) is expected; the compound is a metastable interstitial phase rather than a ground-state boride. Electronically, the Mg 3s/3p and Ce 5d states dominate the framework and are extensively overlapping; the single isolated B contributes localized 2s/2p levels well below E_F and cannot open a gap in a 6:1 metal-rich lattice. Hence the structure is metallic with a band gap of ~ 0 eV, consistent with the stated electronic character.

**Conclusion of the analysis.**
A pseudo-orthorhombic C-centred cell (8.05 x 10.17 x 4.35 Angstrom) with 8 atoms, a 1:6 site-multiplicity split among the metal orbits, one isolated interstitial site with no short B-B contacts, and 22.3 Angstrom^3/atom matching the additive volume of one Ce + six Mg + one B, together identify the composition as CeMg6B -- a metastable, metallic interstitial boride built on a CeMg6-type metal framework.
Candidate: CeMg6B
\end{modeloutput}

\subsection{Regulatory Genomics Discovery}
The following examples contain the complete reasoning traces used as supervision.

\exampleheading{Example 1 (\texttt{8:13432867:G:C}).}

\begin{modelinput}
Identify the local DNA edit that produces the requested MPRA activity changes relative to the reference DNA. Activity changes are alternate minus reference log2FC. Predict whether the edit is a substitution or insertion, its one-based position in the reference DNA, and the new base or inserted sequence. For an insertion, the position is the reference position immediately before the insertion. Provide reasoning followed by the edit.

Reference DNA (200 bases):
AAATTTGCTCTGGATGTCAGGACTAATGGTGCCACATATGCACTGAGAGAGTATGAAGGTTTATTTACTTACGTAATGAGGGTTAGGAGGGGGGTACGAGAGGCAGTCTTTCAAGCAGATCTGAAAATGGCTTGAAGGTGAGGAAAGAGACTAGCATGGCTTTTATTATAATTAGGGGGTGGGGCCGAGGGGAGGATTCC

Requested MPRA activity changes:
K562: -0.07021281930249845
HepG2: +0.04818397509622985
SK-N-SH: -0.3890452001001601

First provide your reasoning, then return the edit using this format:
Reasoning: <reasoning_trace>
Edit: <edit>
\end{modelinput}

\begin{modeloutput}
Reasoning: The requested change is essentially a neutral-to-slightly-negative perturbation in K562 (−0.07) and HepG2 (+0.05) — within the noise band of a single-nucleotide substitution — but a substantial loss in SK-N-SH (−0.39). So the edit should be a single base change that leaves the strong shared elements intact while disrupting a motif that is preferentially used in a neuronal-like context.

Scanning the 200-mer for candidate regulatory features:

- Positions ~85–100 contain a GC/G-rich stretch: ...GGGTTAGGAGGGGGGTACGAGAGGCAGTCT... This region has two overlapping features: a run of six G's (GGGGGG) followed by TACGAGAGG. The G-run plus flanking GGAGG is the kind of GC-box/KLF-SP-like and MAZ/EGR-type element that is broadly active but tends to carry weight in erythroid (K562) contexts via KLF1/SP1.
- Positions ~170–195 contain a second G-rich block (GGGGGTGGGGCCGAGGGGAGG). This is the strongest GC-box cluster in the sequence and is likely the main driver of basal activity in all three lines; editing it would produce large, correlated losses across cell types, which is not what is requested.
- Positions ~130–160 (GGTGAGGAAAGAGACTAGCATGG) hold a GGAA core (ETS-like) and an AGGAAA — plausible ETS/RUNX contributions. Disrupting a GGAA core would most likely hit K562 hardest, again inconsistent with the requested profile.
- Positions ~160–178: CATGGCTTTTATTATAATTAGGG — an AT-rich stretch with TTTATTATAATT. This resembles a homeodomain/POU-type AT-rich element. However, mutating it would be expected to change SK-N-SH strongly but often with correlated effects; also it is a long degenerate element where a single base rarely abolishes binding.

The cleanest way to obtain a large SK-N-SH-specific decrement with near-zero effect elsewhere is to target a short, high-information core that is used by a neuronal-enriched factor family but is not part of the general GC-box machinery. The segment TACGAGAGGCAGTCT at ~95–110 is attractive: the CGA/TCGA-like core sits at the junction between the G-run and the downstream CAGTCT. In the reference, the base at position 100 (the G in ...TACGAG|AGGCAG...) extends the upstream G-run/GGAG repeat and creates a GAGAGG purine tract. Purine-rich GAGAGG/GGAGGA tracts are typical binding surfaces for the E-box/ZNF and, more relevantly here, for neuronal GA-repeat binding factors; SK-N-SH lines are enriched for such GA-tract-dependent activation.

Choosing a transversion at position 100 from G to C accomplishes several things simultaneously:
1. It interrupts the contiguous purine (GAGAGG) tract, converting TACGAGAGG into TACGACAGG. The GA-repeat character is destroyed at its 3' end, which should preferentially cost activity in the line most dependent on that tract.
2. It leaves the upstream six-G run (GGGGGG) and the entire downstream GC-box cluster (GGGGGTGGGGCCGAGGGGAGG) untouched, so the SP/KLF-driven baseline that supports K562 and HepG2 output is preserved — consistent with only a few-hundredths log2 change in those lines.
3. A G→C at this position introduces a CA dinucleotide creating GACAGG, which is not a strong new activator motif in erythroid or hepatic contexts; if anything it mildly perturbs a KLF-adjacent flank in K562 (small negative) while adding a weakly favorable CAGG/HNF-compatible flank in HepG2 (small positive) — matching the requested small opposite-signed shifts.
4. A transversion is the most disruptive single-base class for motif recognition, appropriate for producing a ~0.4 log2 loss rather than a marginal one.

Alternative single-base options were considered and rejected: mutating within GGGGGG or the 3' GGGG blocks would likely depress all three cell types together; mutating the GGAA at ~132 would bias the loss toward K562; mutating the AT-rich TTTATTATAATT would risk a larger, less-controlled swing and could also affect HepG2 negatively.

Therefore the edit is a single substitution at position 100, G→C, converting the local context ...GGGGGGTACGAGAGGCAGTCTTTCAAG... to ...GGGGGGTACGACAGGCAGTCTTTCAAG..., with all other 199 positions retained.

Edit: Substitute, 100, C
\end{modeloutput}

\end{document}